\documentclass{article}

\usepackage{arxiv}
\usepackage{array}

\usepackage[utf8]{inputenc} 
\usepackage[T1]{fontenc}    
\usepackage{hyperref}       
\usepackage{url}            
\usepackage{booktabs}       
\usepackage{amsfonts}       
\usepackage{nicefrac}       
\usepackage{microtype}      
\usepackage{lipsum}
\usepackage{graphicx}
\usepackage{amsmath}
\usepackage{algpseudocode}
\usepackage{algorithm}
\usepackage{pbox}
\usepackage{mathtools}
\usepackage{adjustbox}
\usepackage[numbers]{natbib}
\usepackage[super]{nth}
\usepackage{url}
\usepackage{subfigure}
\usepackage{multirow}
\usepackage{interval}
\usepackage{multirow}
\usepackage{array,multirow}
\usepackage{booktabs, tabularx}
\usepackage{capt-of}
\usepackage{caption} 
\usepackage{booktabs, tabularx}
\usepackage{float}
\usepackage{stfloats}
\usepackage[normalem]{ulem}
\useunder{\uline}{\ul}{}

\usepackage{nicematrix}
\title{Deep Multilabel CNN for Forensic Footwear Impression Descriptor Identification}

\author{
Marcin Budka, Akanda Wahid -Ul- Ashraf, Matthew Bennett\\
    Faculty of Science and Technology\\
    Bournemouth University\\
    Fern Barrow\\
    Poole BH12~5BB, UK\\
    e-mail: \texttt{\{aashraf, mbudka, mbennett\}@bournemouth.ac.uk}\\
    \And
Scott Neville, Alun Mackrill\\
    Bluestar Software Ltd.\\ Fair Cross Offices\\ Stratfield Saye RG7 2BT, \\ {e-mail: \texttt{\{scott.neville, alun.mackrill\}@bluestar-software.co.uk}}
}
\begin{document}
\maketitle

\begin{abstract}
In recent years deep neural networks have become the workhorse of computer vision. In this paper, we employ a deep learning approach to classify footwear impression's features known as \emph{descriptors} for forensic use cases. Within this process, we develop and evaluate an effective technique for feeding downsampled greyscale impressions to a neural network pre-trained on data from a different domain. Our approach relies on learnable preprocessing layer paired with multiple interpolation methods used in parallel. We empirically show that this technique outperforms using a single type of interpolated image without learnable preprocessing, and can help to avoid the computational penalty related to using high resolution inputs, by making more efficient use of the low resolution inputs. We also investigate the effect of preserving the aspect ratio of the inputs, which leads to considerable boost in accuracy without increasing the computational budget with respect to squished rectangular images. Finally, we formulate a set of best practices for transfer learning with greyscale inputs, potentially widely applicable in computer vision tasks ranging from footwear impression classification to medical imaging.
\end{abstract}


\section{Introduction}
\label{sec:Introduction}
In this work we develop an approach to train a deep Convolutional Neural Network (CNN) to classify features in footwear impressions for use in forensic applications. The features we classify are known as \emph{descriptors} within the UK footwear forensic units~\cite{bluestardesciptor, milne2012forensic, hannah-montana-2020, bennett2018digital} and can be defined as recognisable units within a footwear pattern which can be classified. The \emph{descriptors} are used by forensic practitioners to describe the makeup of a footwear pattern.

Every footwear impression added to the UK’s National Footwear Reference Collection (NFRC)~\footnote{The National Footwear Reference Collection (NFRC) and The National Footwear Database (NFD) are developed and maintained by Bluestar Software Ltd (BSL)~\cite{bluestardesciptor}} is manually labelled with the \emph{descriptors}~\cite{bluestardesciptor}. The NFRC is built on an agreed standard for coding footwear patterns for different forces in the UK and at the time of writing, to the best of our knowledge, is the biggest police-owned collection of footwear impressions in the world. The NFRC footwear pattern collection is updated on a regular basis~\cite{bluestarnfdnfrc}.

The NFD is a successor of the NFRC where footwear labels are maintained and added regularly. The NFRC records the custody and crime scene marks while the NFD facilitates matching with the NFRC footwear patterns. Currently, around 30 out of 43 police forces in England and Wales, continuously send or update data in the NFD~\cite{bluestarnfdnfrc}.  

The NFRC uses a total of 17 \emph{descriptors} to identify a footwear impression. Each of the \emph{descriptors} is assigned a unique name and code. A shoe print or footwear impression may contain any subset of these \emph{descriptors}. The location of the \emph{descriptors} are divided into two parts: 1)~the heel / instep, and 2)~the main sole (i.e. top). In this study we do not exploit this location information in any way. A single \emph{descriptor} can exist multiple times in a shoe print, however, the specific location (other than the heal/instep or main sole) and frequency of the \emph{descriptor} is not identified and counted.

Each of the 17 \emph{descriptors} (Table~\ref{tab:desciptor_code_names}) has specific semantics (for the purpose of quick identification by forensics practitioner rather than a computer), which relate to the name of the \emph{descriptor}. For example, \emph{descriptor} \emph{D05: 5 sided}, contains all shapes which are 5 sided; \emph{descriptor} \emph{D09: Text} indicates any text that can be found on a shoe print. The number of possible geometric variations that are usually found can potentially be infinite. For example, \emph{descriptor} \emph{D09: Text} can be any combination of characters and fonts, while \emph{descriptor} \emph{D05: 5 sided} can be a rough pentagon of any shape and form. Two \emph{descriptors} can overlap, resulting in a multiple \emph{descriptors} from a single topological subpattern on a footwear impression. For example, \emph{descriptor} \emph{D09: Text} and \emph{D10: Logo} usually overlap, as many logos contain text. Additionally, among the 17 \emph{descriptors}, three are subcategories of two main/parent \emph{descriptors}: \emph{D01-01: Wavy}, \emph{D01-D02: Curved-wavy} are the subcategories of \emph{D01: Bar}, and \emph{D02-01: Target} is a single subcategory of \emph{D02: Circular}. While labelling with the \emph{descriptors} for a footwear impression, the microscopic patterns of the impressions are not usually considered. For example, \emph{D12: Texture} can contain microscopic patterns which are also \emph{D06: 6 sided} but usually \emph{D06: 6 sided} is not labelled in such cases as these microscopic patterns are often not reliable and persistent~\cite{bodziak1999footwear}. All the sided shaped \emph{descriptors} (e.g. \textit{D03}, \textit{D04}, etc.) do not necessarily have very precise straight lines as sides but some curves and deformations are ubiquitous.

\begin{table}[]
\scriptsize
\begin{center}
\begin{adjustbox}{width=0.8\linewidth}
\begin{tabular}{|l|l|l|l|}
\hline

\multirow{2}{*}{\textbf{\begin{tabular}[c]{@{}l@{}}D01\\ Bar\end{tabular}}} & 
\begin{tabular}[c]{@{}l@{}}A bar of any type such as straight, \\ angled, curved, including chevrons\end{tabular} & \multirow{2}{*}{\textbf{\begin{tabular}[c]{@{}l@{}}D07\\ Complex\end{tabular}}} & 
\begin{tabular}[c]{@{}l@{}}Shapes such as star, arrow, waisted \\ bar, heart and cross, and any other \\ shape with more than six sides\end{tabular} \\ &
\includegraphics[width=40mm,height=10mm]{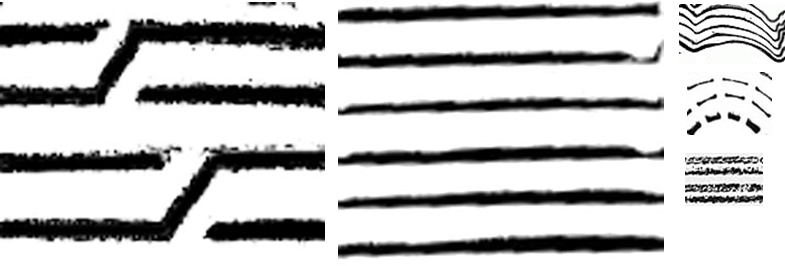} & &
\includegraphics[width=40mm,height=10mm]{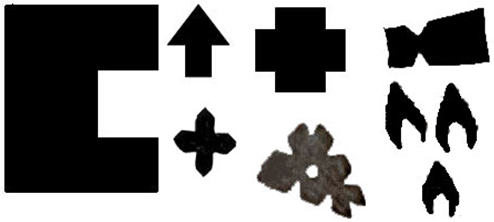} \\ \hline

\multirow{2}{*}{\textbf{\begin{tabular}[c]{@{}l@{}}D01-01\\ Wavy\end{tabular}}} & 
\begin{tabular}[c]{@{}l@{}}A bar element with more than one \\ directional change\end{tabular} & \multirow{2}{*}{\textbf{\begin{tabular}[c]{@{}l@{}}D08\\ Zigzag\end{tabular}}} & 
\begin{tabular}[c]{@{}l@{}}A broken or continuous line that \\ changes direction repeatedly \\ with abrupt right and left turns\end{tabular} \\ &
\includegraphics[width=40mm,height=10mm]{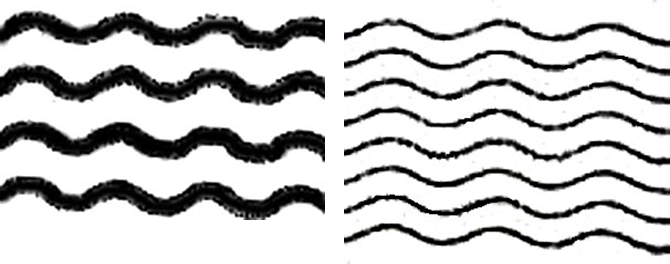} & &
\includegraphics[width=40mm,height=10mm]{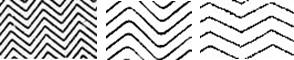} \\ \hline

\multirow{2}{*}{\textbf{\begin{tabular}[c]{@{}l@{}}D01-02\\ Curved-\\wavy\end{tabular}}} & 
\begin{tabular}[c]{@{}l@{}}Any bar shape/ element deviating \\ from a straight line with a single \\ rounded directional change however \\ small the angle of the curved section\end{tabular} & 
\multirow{2}{*}{\textbf{\begin{tabular}[c]{@{}l@{}}D09\\ Text\end{tabular}}} & 
\begin{tabular}[c]{@{}l@{}}Any alpha-numeric characters; \\ may overlap with D10\end{tabular}  \\ & \includegraphics[width=40mm,height=10mm]{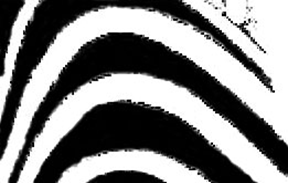} & &
\includegraphics[width=40mm,height=15mm]{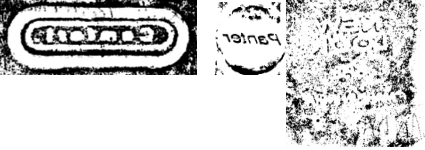} \\ \hline

\multirow{2}{*}{\textbf{\begin{tabular}[c]{@{}l@{}}D02\\ Circular\end{tabular}}} & 
\begin{tabular}[c]{@{}l@{}}Includes circle, semi-circle, oval, \\ semi-oval, concentric circles, \\ target, tear-drop, stud,  crescent\end{tabular}
& \multirow{2}{*}{\textbf{\begin{tabular}[c]{@{}l@{}}D10\\ Logo\end{tabular}}} & 
\begin{tabular}[c]{@{}l@{}}A brand or trademark incorporating \\ a symbol, badge, emblem or picture;\\ may  overlap with D09\end{tabular}  \\
& \includegraphics[width=40mm, height=10mm]{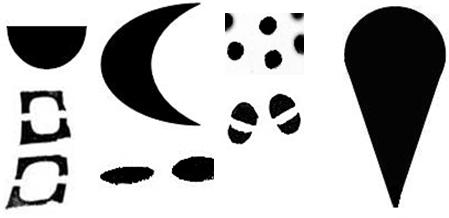} &                                                & \includegraphics[width=40mm, height=10mm]{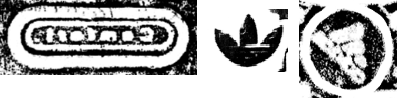} \\ \hline
                                                                            
\multirow{2}{*}{\textbf{\begin{tabular}[c]{@{}l@{}}D02-01\\ Target\end{tabular}}} & 
\begin{tabular}[c]{@{}l@{}}Any concentric circle arrangement \\ whether the centre-most circle is \\ hollow or solid\end{tabular}
& \multirow{2}{*}{\textbf{\begin{tabular}[c]{@{}l@{}}D11\\ Lattice\end{tabular}}} & 
\begin{tabular}[c]{@{}l@{}}A regular, interlocking and/or repeated \\ pattern (aka network, web or trellis);\\ includes brickwork, herring-bone, \\ honeycomb and chicken wire\end{tabular}  \\
& \includegraphics[width=40mm,height=10mm]{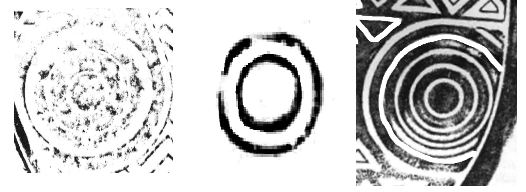} &                                                & \includegraphics[width=40mm,height=10mm]{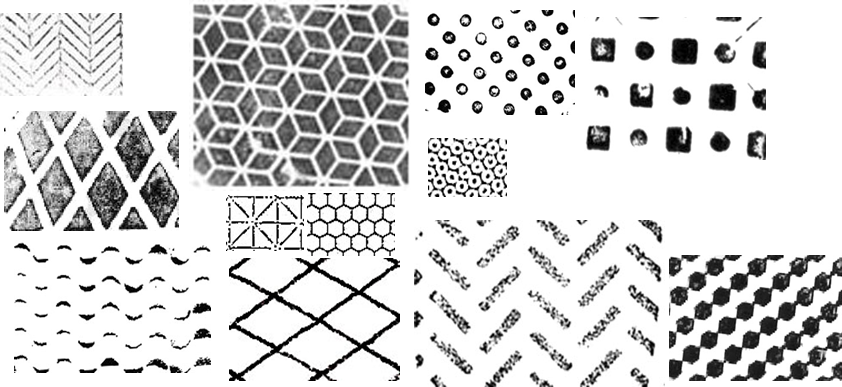} \\ \hline

\multirow{2}{*}{\textbf{\begin{tabular}[c]{@{}l@{}}D03\\ 3 sided\end{tabular}}} & 
\begin{tabular}[c]{@{}l@{}}All types of triangle including those \\ with one rounded side such as a \\ pie-segment\end{tabular} 
& \multirow{2}{*}{\textbf{\begin{tabular}[c]{@{}l@{}}D12\\ Textured\end{tabular}}} & 
\begin{tabular}[c]{@{}l@{}}This includes pre-dominant stippling, \\ crepe or random patterns added by \\ the manufacturer as part of  their design\end{tabular}  \\
& \includegraphics[width=40mm, height=10mm]{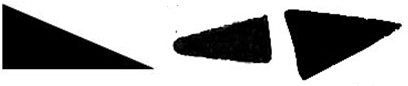} &                                                & \includegraphics[width=40mm, height=10mm]{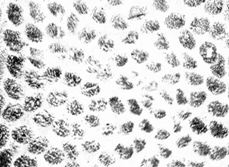} \\ \hline

\multirow{2}{*}{\textbf{\begin{tabular}[c]{@{}l@{}}D04\\ 4 sided\end{tabular}}} & 
\begin{tabular}[c]{@{}l@{}}Square, rectangle, oblong, paralle-\\logram, rhombus, diamond, arrowhead\end{tabular} 
& \multirow{2}{*}{\textbf{\begin{tabular}[c]{@{}l@{}}D13\\ Hollow\end{tabular}}} & 
\begin{tabular}[c]{@{}l@{}}A pattern that has the appearance of a \\ hollow shape, such as a doughnut or frame\end{tabular}  \\
& \includegraphics[width=40mm, height=10mm]{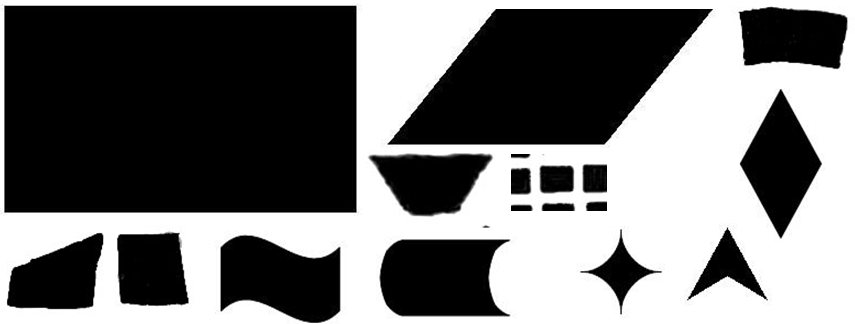} &                                                & \includegraphics[width=40mm, height=10mm]{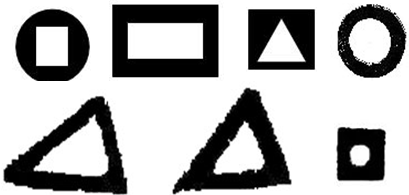} \\ \hline

\multirow{2}{*}{\textbf{\begin{tabular}[c]{@{}l@{}}D05\\ 5 sided\end{tabular}}} & 
\begin{tabular}[c]{@{}l@{}}Usually a regular shaped pentagon, \\ but includes all five-sided shapes\end{tabular} 
& \multirow{2}{*}{\textbf{\begin{tabular}[c]{@{}l@{}}D14\\ Plain\end{tabular}}} & 
\begin{tabular}[c]{@{}l@{}}A plain surface with no patterns \\ or texture\end{tabular}  \\
& \includegraphics[width=40mm, height=10mm]{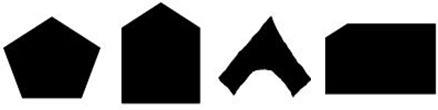} &                                                & \includegraphics[width=40mm, height=10mm]{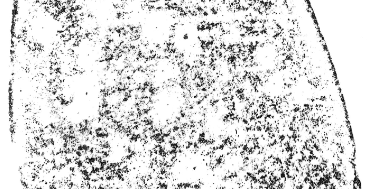} \\ \hline

\multirow{2}{*}{\textbf{\begin{tabular}[c]{@{}l@{}}D06\\ 6 sided\end{tabular}}} & 
\begin{tabular}[c]{@{}l@{}}Usually a regular shaped hexagon, \\ but includes all six-sided shapes\end{tabular} & \multicolumn{2}{l}{\multirow{2}{*}{\textbf{}}} \\ & 
\includegraphics[width=40mm, height=10mm]{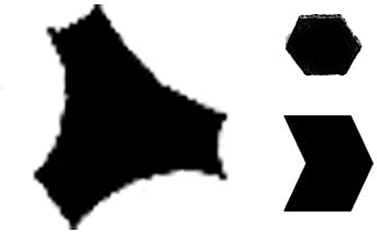} & 
\multicolumn{2}{l}{}                           \\ \cline{1-2}

\end{tabular}
\end{adjustbox}
\end{center}
\caption{Footwear \emph{descriptors} for the UK’s National Footwear Reference Collection (NFRC)}
\label{tab:desciptor_code_names}
\end{table}


\subsection{Footwear Impression Imaging Methods}
\label{sec:imaging_techniques}
In UK policing collection of footwear evidence is normally done in two scenarios: 1)~collection of detainee footwear in custody, and 2)~collection of crime scene marks. The vast majority of the footwear impressions captured from detainees in custody usually follow one of the below processes~\cite{footwearmarksrecoveryman}:
\begin{itemize}
    \item \textit{Inked Impressions:} The inked impressions are captured using a specialist pad and paper kit (sometimes called a `Bigfoot Kit')~\cite{bluestardesciptor,bodziak1999footwear,milne2012forensic}. The kit uses a pad with a reactive chemical and specialist paper. The impressions can then be digitised using an office document scanner, if required. 
    \item \textit{Ink-less Impressions:} A specialised footwear impression digital scanner is used in this case to capture the footwear impression without any use of ink. This process produces only a digital copy of the impression whereas the inked impression also produces a physical copy on paper~\cite{bluestardesciptor}.
\end{itemize}

Additionally, some UK forces use coloured photographs of the shoe sole as opposed to using one of the impression capturing methods described above~\cite{bluestardesciptor}.

\subsection{Identifying Descriptors}
\label{Identifying Descriptors}
In practice to date, the \emph{descriptors} are manually identified by experts and are only used as an intermediate step to identify a pattern. Processes vary between police forces, however when adding an impression to the NFRC, two independent experts individually identify the \emph{descriptors}. If both experts agree on the set of identified \emph{descriptors}, the footwear impression image is labelled with the identified \emph{descriptors}. However, when there is a disagreement between the two experts, the labelling process involves a panel of experts for further analysis. The accuracy of identifying the \emph{descriptors} by experts are thought to be `very high', however, to the best of our knowledge, there was no empirical study to quantify this accuracy~\cite{bluestardesciptor}.

\subsection{Limitations}
The main limitation associated with manually identifying \emph{descriptors} is the time and cost of human expertise. Although forensic practitioners are able to directly identify many common footwear impressions without the need for classification against the \emph{descriptors}, classifying rare or new shoe models takes longer. As there are tens of thousands of footwear models, it is impractical for a human expert to be able to accurately identify a specific model with only the \emph{descriptors}. The NFRC/NFD provides a number of additional searching and ordering features to make identification possible in a practical time span. These features are generally used in the same way for all searches (looking at frequency/geography of distribution) and therefore take little time to use compared with the time taken to identify descriptors.
However, the most frequently worn footwear are very well known to the forensic practitioners thus are easily labelled by them, without the need of any computer system, or the \emph{descriptors}. 
Since the accuracy of human footwear forensics experts are not empirically evaluated, the automated process cannot be argued to be same or better than human experts. Despite this, clear use cases for an automatic \emph{descriptor} identification exist. For example, when a new footwear model is captured, labelling would be completed by an expert, then blindly verified by another. An automatic \emph{descriptor} identification will be faster and have higher availability for the second check as the number of human experts available is limited. Automatic \emph{descriptor} identification could potentially replace the second opinion when adding patterns to the NFRC (see Section~\ref{Identifying Descriptors}). 


\section{Automated Descriptor Inference}
\label{An Approach to Automate Descriptor Inference}
\label{Use-case of an Automated Descriptor Inference}
The automation of the \emph{descriptor} analysis can provide rapid identification of the \emph{descriptors} in a given impression, which in turn will result in faster identification of a shoe model from its print, especially for an untrained (in terms of footwear analysis) personnel. Additionally, the identified \emph{descriptors} can be used to narrow down the search in the database with thousands of footwear impressions. The automated approach, which is not only capable of identifying the \emph{descriptors} but also infer their topological location (e.g. using Grad-CAM~\cite{selvaraju2017grad}) can be further beneficial for training police users. Rapid automatic \emph{descriptor} identification can be achieved without involving a forensic expert, resulting in faster determination of intelligence. The latter is particularly important as the suspect will then have little time to destroy the evidence and can be questioned sooner (ideally before leaving custody), resulting in a plausibly increased detection rate. In England and Wales, there are around 25-30 (an estimation without an official source) human experts who can identify the \emph{descriptors} currently, whereas there are 123,171\footnote{According to a statistical bulletin published on the 18\textsuperscript{th} July 2019 by the Home Office for England and Wales. This number of officers does not include the British Transport Police} law enforcement personnel~\cite{HomeOfficePolice_numbers} who may handle a case where identification of the \emph{descriptors} may be necessary. Automated \emph{descriptor} identification can potentially provide such expertise to all the law enforcement personnel in the UK. 

Due to large variability in the complex geometric shapes and patterns of a \emph{descriptor}, a simple template matching algorithm~\cite{brunelli2009template} would be suboptimal. Each of the \emph{descriptors} has an apparent but variable high-level geometric semantics.

\begin{figure}[htp]
    \centering
    \includegraphics[width=12cm]{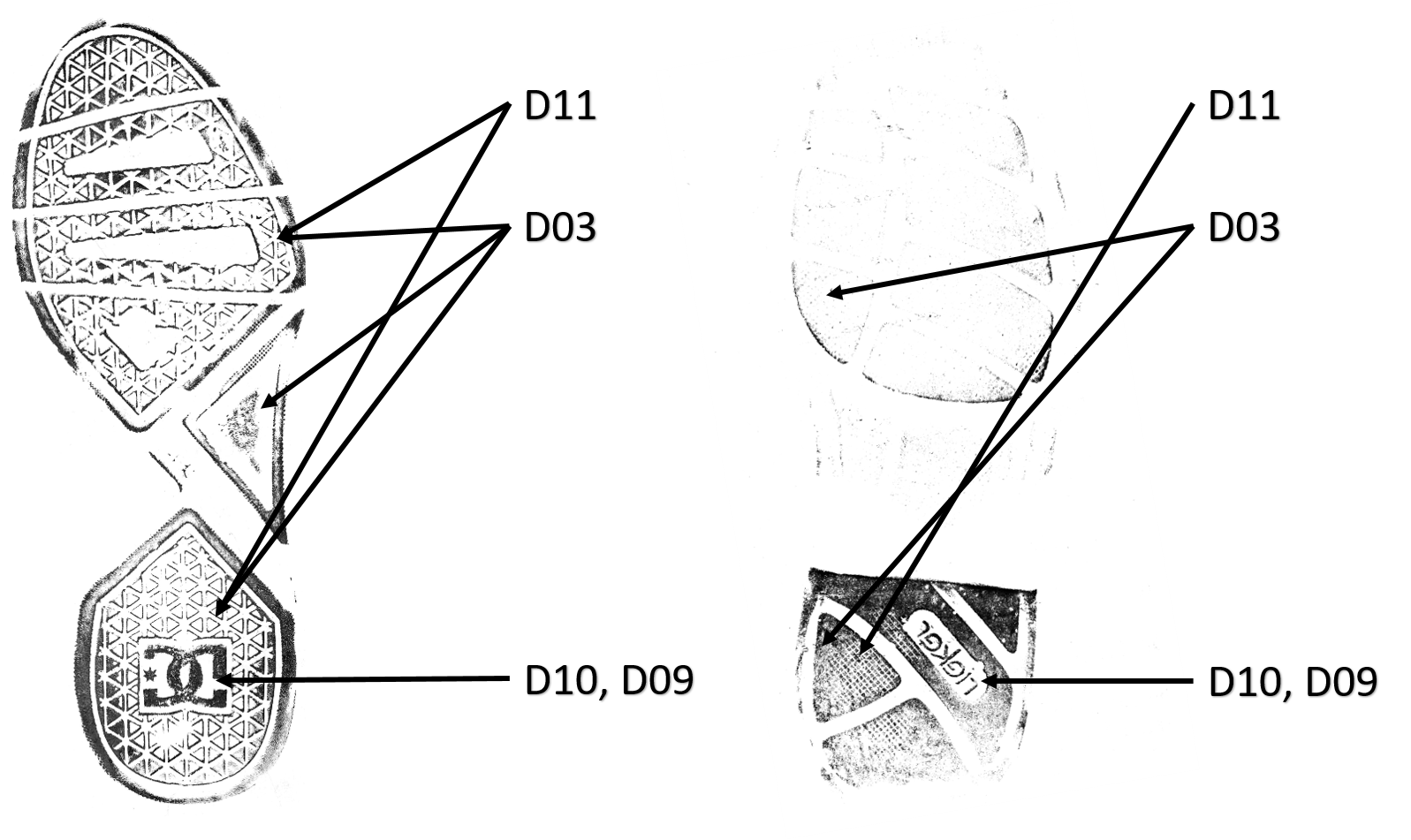}
    \caption{Different types of~\emph{descriptors}: D10, D11, D03 on two separate real inked impressions}
    \label{fig:conv_net}
\end{figure}

Figure~\ref{fig:conv_net} shows two real-world inked impression with four \emph{descriptors} each; \textit{D03}, \textit{D09}, \textit{D10}, \textit{D11}. As it can be seen, although the same descriptors appear on both of the impressions, their patterns are very distinct. While \textit{D10: Logo} is an obvious example, a more `stable' \textit{D03: 3 sided} also looks quite different. The impression on the right has \textit{D03} with smother edges, and also bigger in size than \textit{D03} found in the left impression. Also note that although both the impressions have \textit{D11: Lattice}, its appearance is very distinct. 

As a result, designing filters to identify the \emph{descriptors} is not practical. Instead, a deep learning based approach has been taken, able to automatically learn the filters from the already existing manually labelled dataset.

\section{Input Image Resolution in Deep Neural Networks}
\label{sec:Image Resolution for Deep Learning Models}
Training a deep neural network requires estimating a large number of parameters in the order of hundreds of millions. The matrix arithmetic operation performed to estimate these parameters are best suited for a Graphics Processing Unit (GPU) due to a GPU's better ability to perform highly parallel floating-point operations when compared with a CPU (Central Processing Unit). The GPU computational power are still limited however and there are other bottlenecks like moving data between the main memory and the GPU.  
As a result, a smaller model with a lower number of parameters is computationally more efficient than a bigger model with a larger number of parameters. 

Apart from the base architecture (number of layers and units per layer) of a neural network, the number of computations grows approximately quadratically with the resolution of the input image. Higher resolution images also take up more space in the GPU memory, which tends to be smaller than system memory, limiting the batch size and further reducing the overall training speed. In order to reduce the computational cost and facilitate faster training, the input images are usually downscaled~\cite{koziarski2018impact, kannojia2018effects}. However, the performance/accuracy of a neural network tends to suffer when image resolution is reduced.

It should also be noted that the theoretical benefit of a higher resolution image may not always increase with an ever increasing resolution of that image, e.g. once we have already achieved the theoretical upper bound of the accuracy for a specific domain. 
In our case, we have very small and complex features defining a class (see Section~\ref{sec:Introduction}), thus, the upper bound of the resolution with a beneficial impact on the model is assumed to be higher than in the classification based tasks where the classes are usually more apparent. 



\section{Image Interpolation Techniques}
\label{sec:Image Interpolation Techniques}
Our dataset consist of high-resolution footwear impressions that have been captured via the means discussed in Section~\ref{sec:imaging_techniques}. As deliberated on in Section~\ref{sec:Image Resolution for Deep Learning Models}, in practice the images resolution need to be reduced and there are a number of different image interpolation~\footnote{We use the terms interpolation, resampling, downscaling, and resizing interchangeably} techniques that can be used here. In our experiments, we investigate and benchmark various combinations of image interpolation techniques, including:

\begin{itemize}
    \item \emph{Nearest Neighbour interpolation (\textbf{N})}, which is the least computationally expensive and does not insert new colours in the result. In this interpolation, only the nearest neighbour's pixel intensity is considered. The estimation function $f$ on a point $(x,y)$ becomes a piecewise function with constant value~\cite{thevenaz2000image,thevenaz2000interpolation}.

    \item \emph{Bilinear interpolation (\textbf{B})} is a linear interpolation over all non-channel dimensions of an image, i.e. for a two dimensional image it is the interpolation over both the $X$ and $Y$ dimensions~\cite{smith1981bilinear}. 
    A straight line passing through two points $(x_1,y_2)$ and $(x_2, y_2)$ between range $x_1$ and $x_2$ is the linear interpolant of these two points. For a range of $(x_1, x_2)$, the slopes of the interpolant from both of these points ($x_1$ and $x_2$) should be exactly the same, hence the following equation of slopes can be formulated:
    
    \begin{equation}
    \frac{y-y_1}{x-x_1} =  \frac{y_2-y_1}{x_2-x_1}
    \label{interpol_1}
    \end{equation}
    
    Solving Equation~\ref{interpol_1} for $y$ gives:  
    \begin{equation}
    y = y_1 \Big( \frac{x_2-x}{x_2-x_1} \Big) +  y_2 \Big( \frac{x-x_1}{x_2-x_1} \Big)
    \label{interpol_2}
    \end{equation}
    Equation~\ref{interpol_2} produces interpolation over the $X$ direction. In case of a two dimensional image for four different points on the image, $Q_{11} = (x_1, y_1)$, $Q_{12} = (x_1, y_2)$,  $Q_{21}=(x_2,y_1)$, $Q_{22}=(x_2, y_2)$, the task is to estimate the function $f$ at a point $(x,y)$. In this four points scenario, the linear interpolation on the $X$ direction using Equation~\ref{interpol_2} gives us the following: 
    
    \begin{equation}
    f(x, y_1) =  \Big( \frac{x_2-x}{x_2-x_1} \Big) f(Q_{11}) + \Big( \frac{x-x_1}{x_2-x_1} \Big) f(Q_{21})
    \label{interpol_y1}
    \end{equation}
    
    \begin{equation}
    f(x, y_2) = \Big( \frac{x_2-x}{x_2-x_1} \Big) f(Q_{12}) +  \Big( \frac{x-x_1}{x_2-x_1} \Big) f(Q_{22})
    \label{interpol_y2}
    \end{equation}
    
    We can then use Equations~\ref{interpol_y1} and~\ref{interpol_y2} to interpolate on the $Y$ direction in order to estimate  $f(x,y)$:
    
    \begin{multline}
    f(x, y) =  \frac{y_2-y}{y_2-y_1} \Bigg( \frac{x_2-x}{x_2-x_1}  f(Q_{11}) + \frac{x-x_1}{x_2-x_1} f(Q_{21})\Bigg)\\  + \frac{y-y_1}{y_2-y_1} \Bigg(  \frac{x_2-x}{x_2-x_1}  f(Q_{12}) +   \frac{x-x_1}{x_2-x_1} f(Q_{22})\Bigg) 
    \label{interpol_x_y}
    \end{multline}
    
    \item \emph{Hamming (\textbf{H})} interpolation technique uses a \emph{sinc} approximating kernel by multiplying (convolution operation as its in the frequency domain) the well-known \emph{sinc}~\cite{woodward1952information} function with the hamming~\cite{blackman1958measurement} window function~\cite{meijering2001quantitative}. Equation~\ref{interpol_hamming_sinc} is the \emph{sinc} function and Equation~\ref{interpol_hamming_window} is the Hamming window function with the window interval $(-m,m)$

    \begin{equation}
    W_{hamming} = 0.54 + 0.46 \cos\bigg( \frac{\pi x}{m} \bigg)
    \label{interpol_hamming_window}
    \end{equation}
    
    \begin{equation}
    sinc(x) = \frac{\sin (\pi x)}{\pi x} 
    \label{interpol_hamming_sinc}
    \end{equation}

\end{itemize}

\begin{figure}[H]
\centering
\includegraphics[width=32mm]{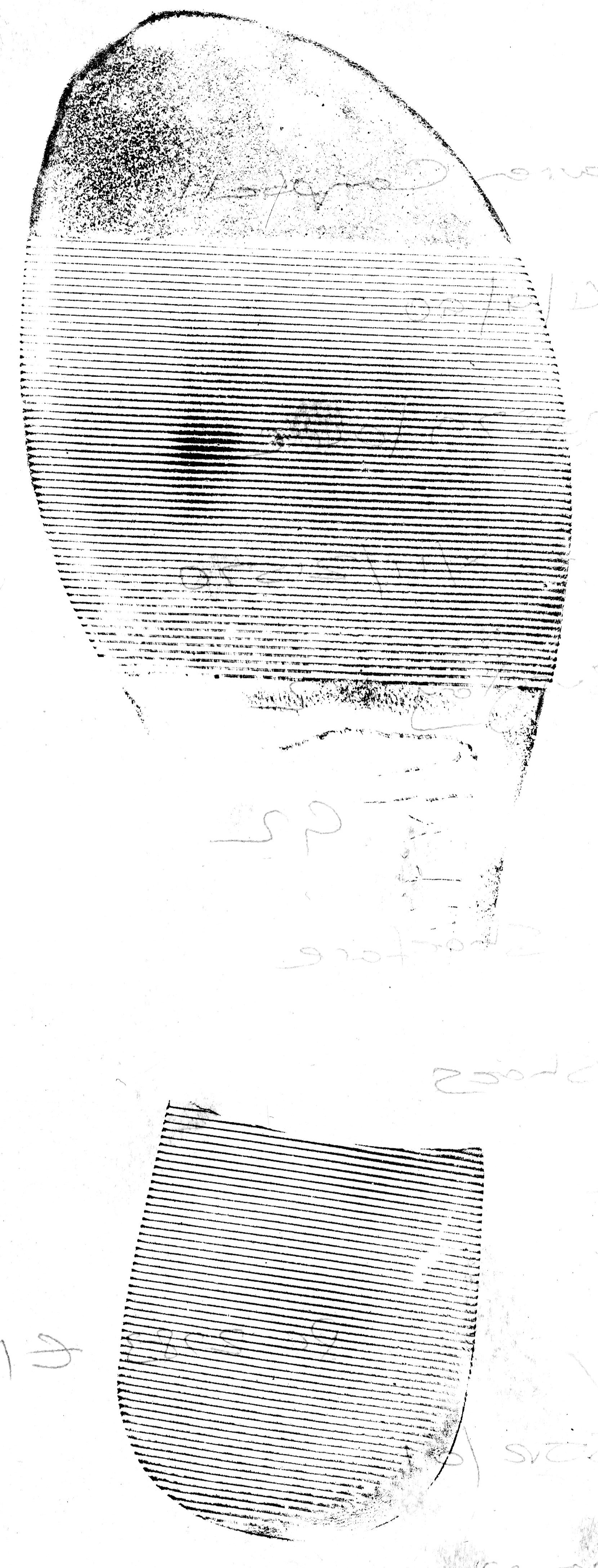}
\caption{Original image without any interpolation. Zoom in to circumvent distortion introduced by the interpolation applied from the medium where this paper is being viewed}
\label{fig:original_no_interpolation}
\end{figure}

\begin{figure}[H]
\centering
\subfigure[$N_{(100 \times 262)}$ ]{\label{fig:n_1}\includegraphics[width=32mm]{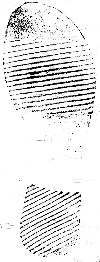}}
\subfigure[$B_{(100 \times 262)}$]{\label{fig:b_1}\includegraphics[width=32mm]{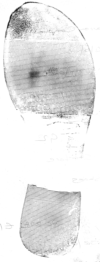}}
\subfigure[$H_{(100 \times 262)}$]{\label{fig:h_1}\includegraphics[width=32mm]{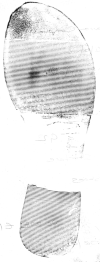}}

\subfigure[$N_{(300 \times 786)}$ ]{\label{fig:n_2}\includegraphics[width=32mm]{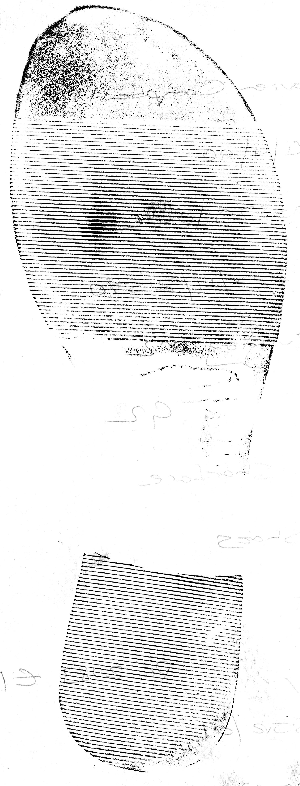}}
\subfigure[$B_{(300 \times 786)}$]{\label{fig:b_2}\includegraphics[width=32mm]{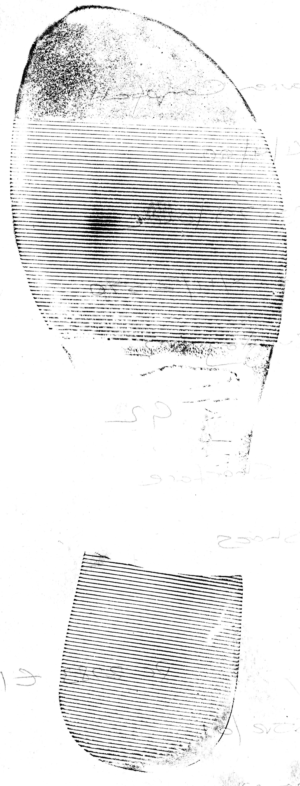}}
\subfigure[$H_{(300 \times 786)}$]{\label{fig:h_2}\includegraphics[width=32mm]{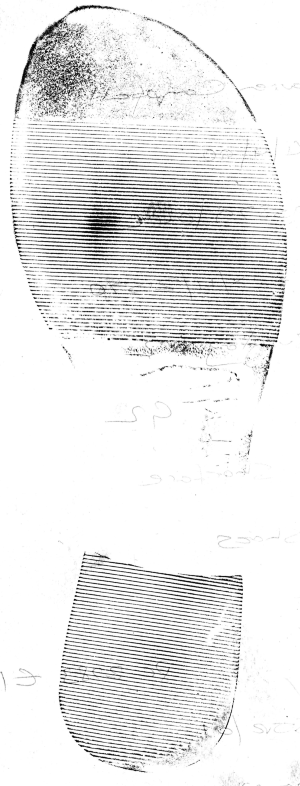}}

\caption{Interpolation samples with fixed aspect ratio and varying sizes}
\label{fig:interpolation_samples_1}
\end{figure}

\begin{figure}[H]
\centering

\subfigure[$N_{(500 \times 1310)}$ ]{\label{fig:n_3}\includegraphics[width=32mm]{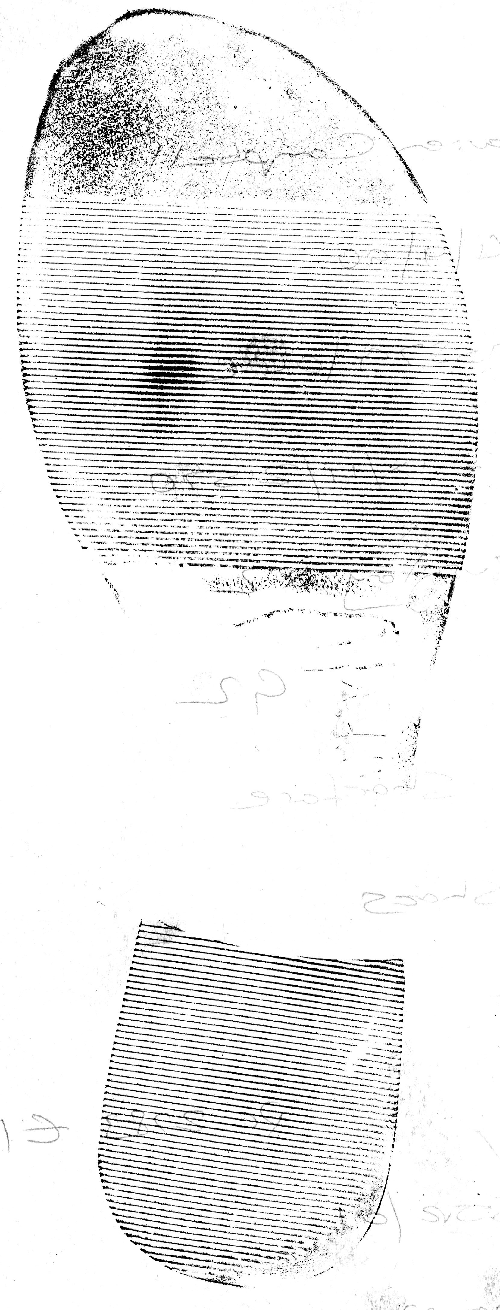}}
\subfigure[$B_{(500 \times 1310)}$]{\label{fig:b_3}\includegraphics[width=32mm]{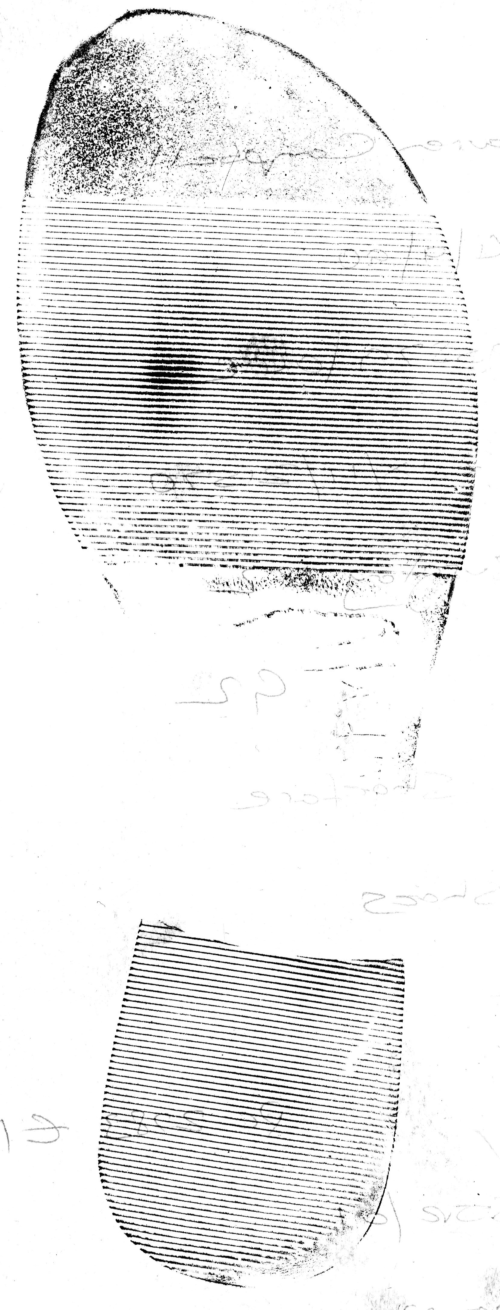}}
\subfigure[$H_{(500 \times 1310)}$]{\label{fig:h_3}\includegraphics[width=32mm]{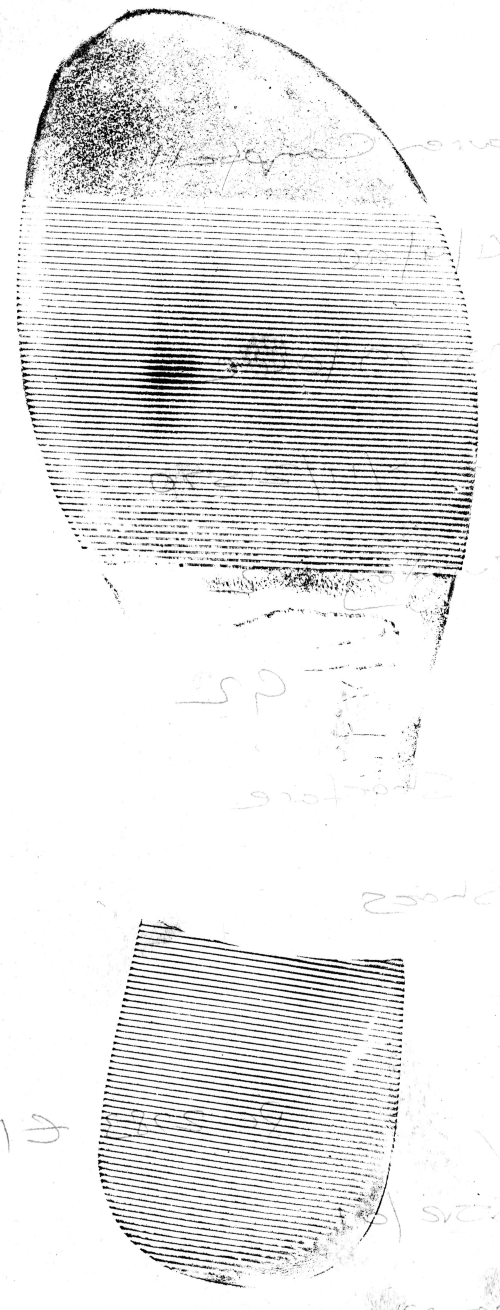}}

\subfigure[$N_{(700 \times 1834)}$ ]{\label{fig:n_4}\includegraphics[width=32mm]{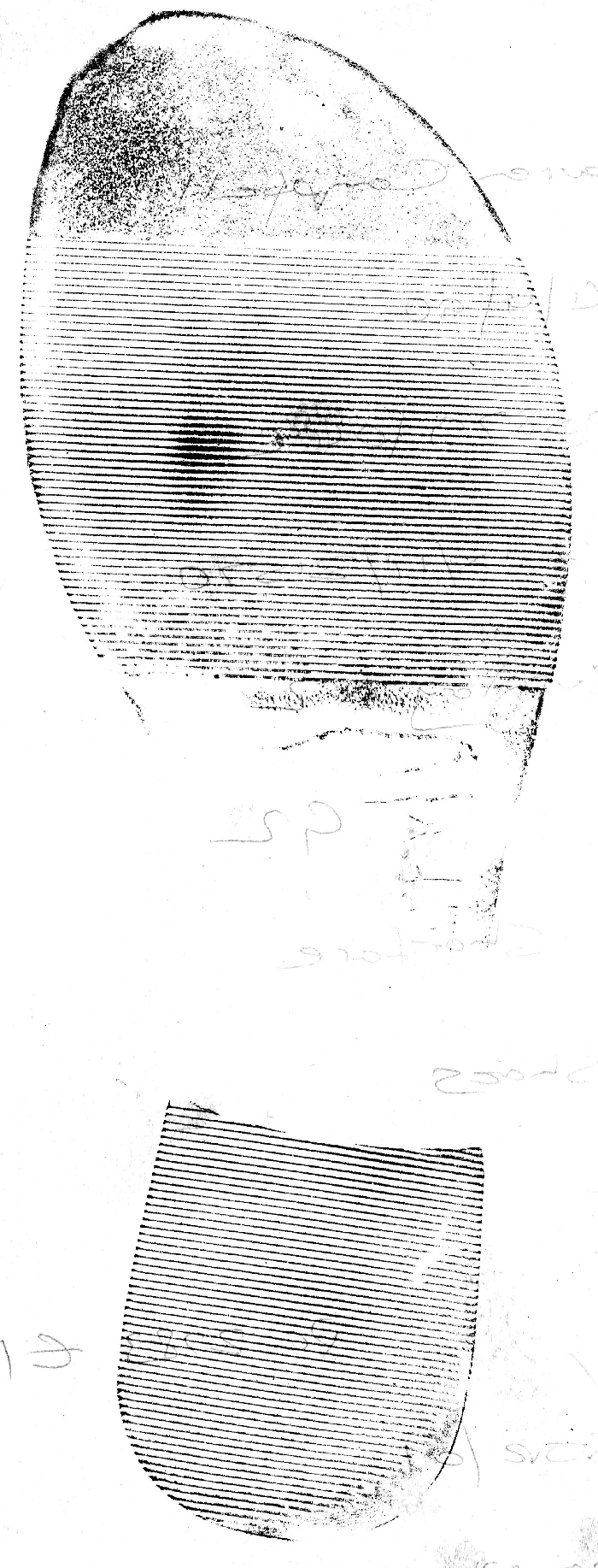}}
\subfigure[$B_{(700 \times 1834)}$]{\label{fig:b_4}\includegraphics[width=32mm]{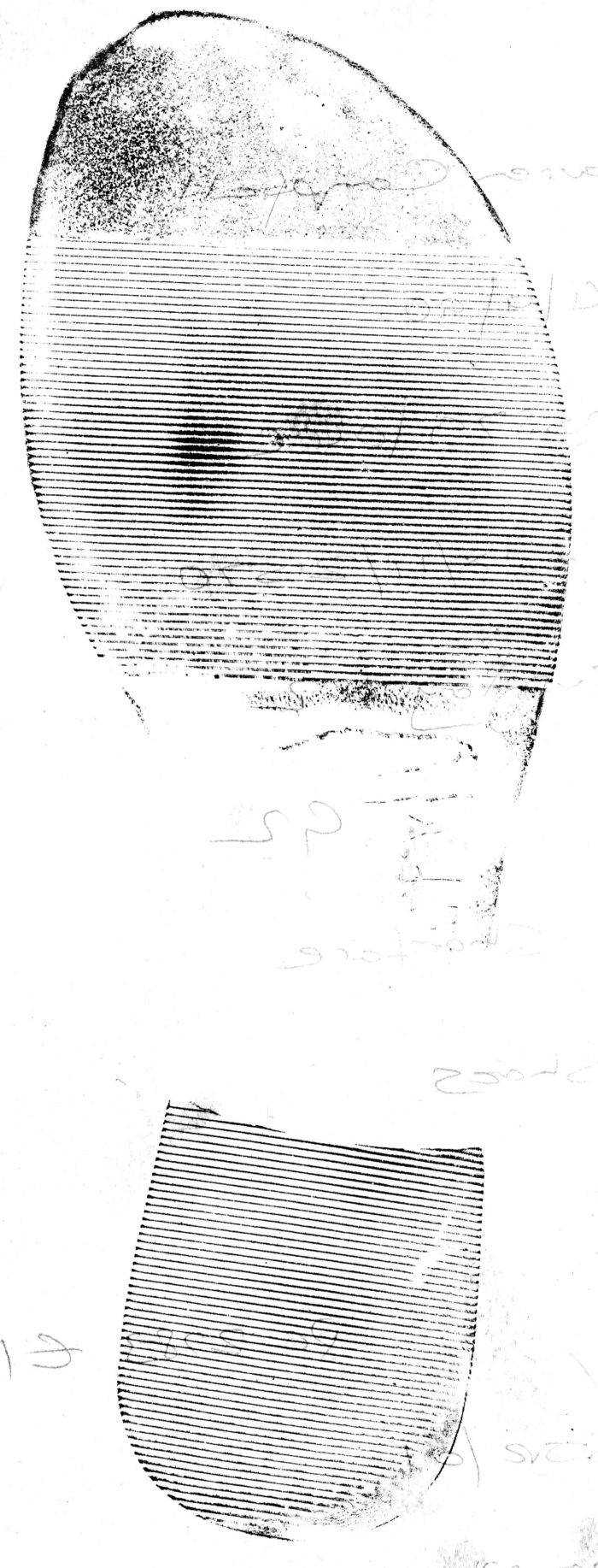}}
\subfigure[$H_{(700 \times 1834)}$]{\label{fig:h_4}\includegraphics[width=32mm]{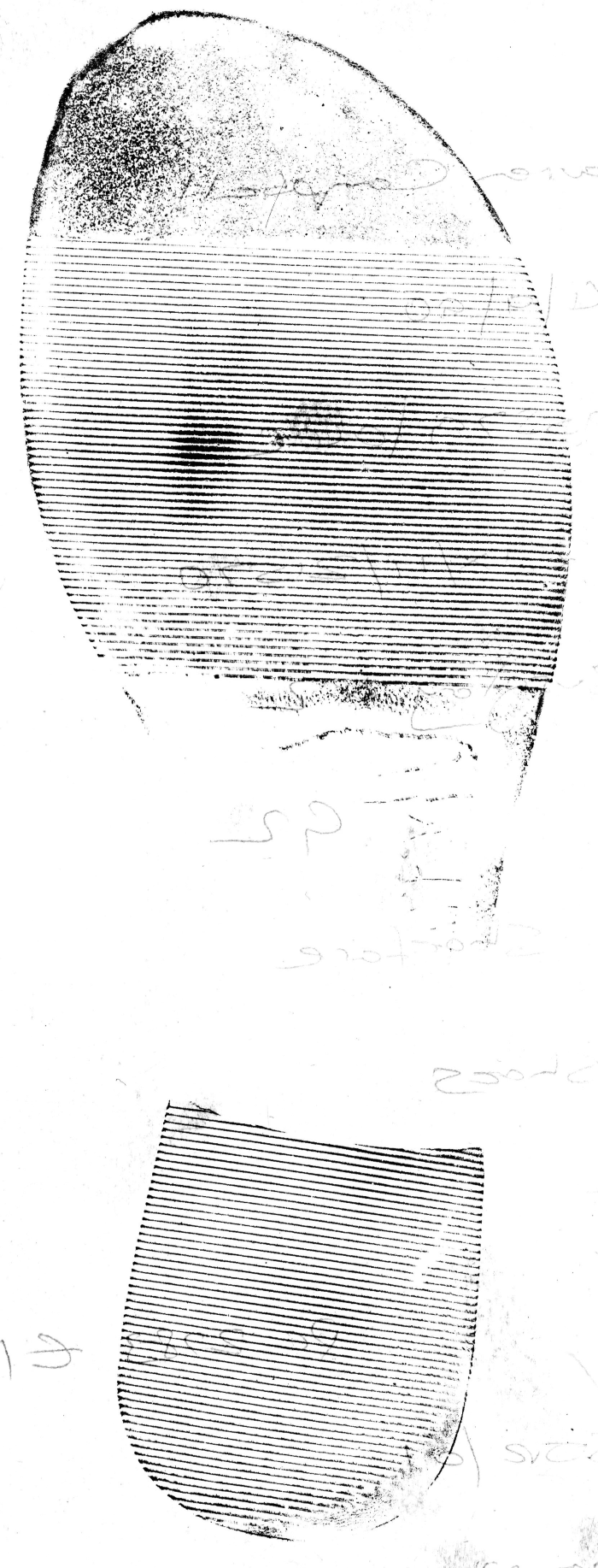}}

\caption{Interpolation samples with fixed aspect ratio and varying sizes (zoom in to see the original pattern)}
\label{fig:interpolation_samples_2}
\end{figure}

\begin{figure}[H]
\centering

\subfigure[$N_{(1000 \times 2620)}$ ]{\label{fig:n_5}\includegraphics[width=32mm]{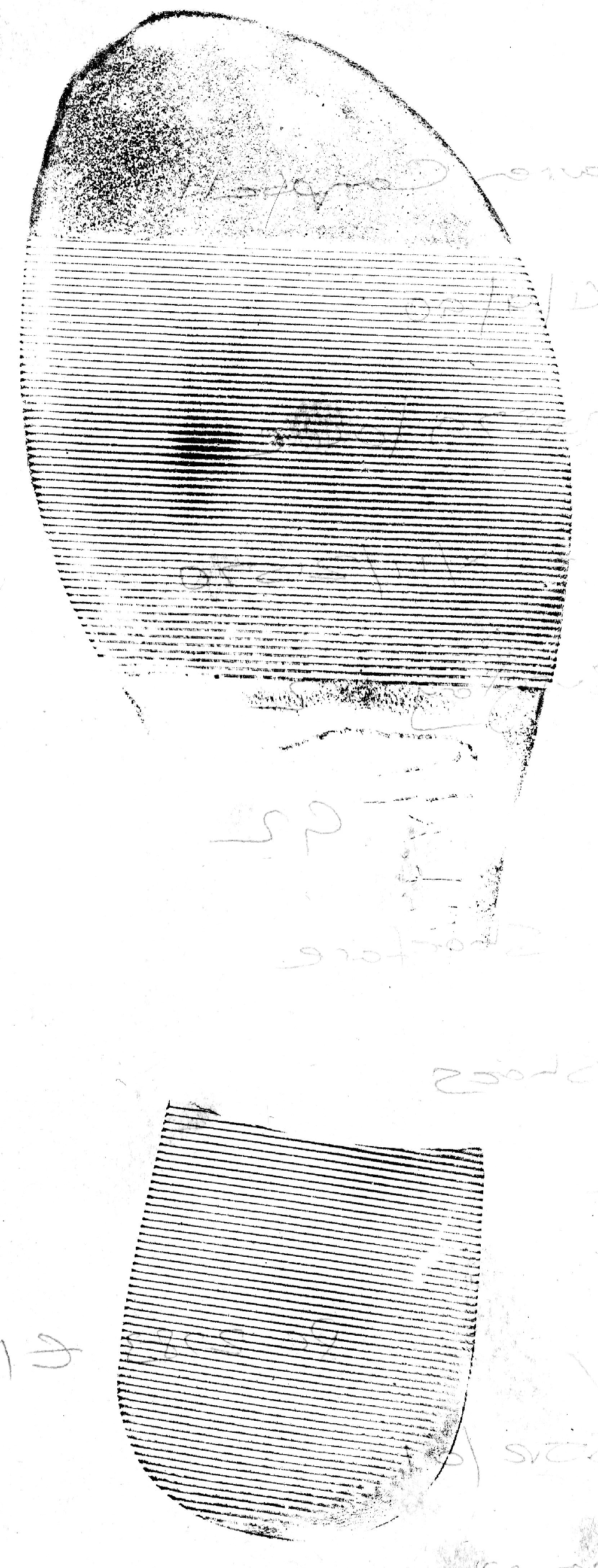}}
\subfigure[$B_{(1000 \times 2620)}$]{\label{fig:b_5}\includegraphics[width=32mm]{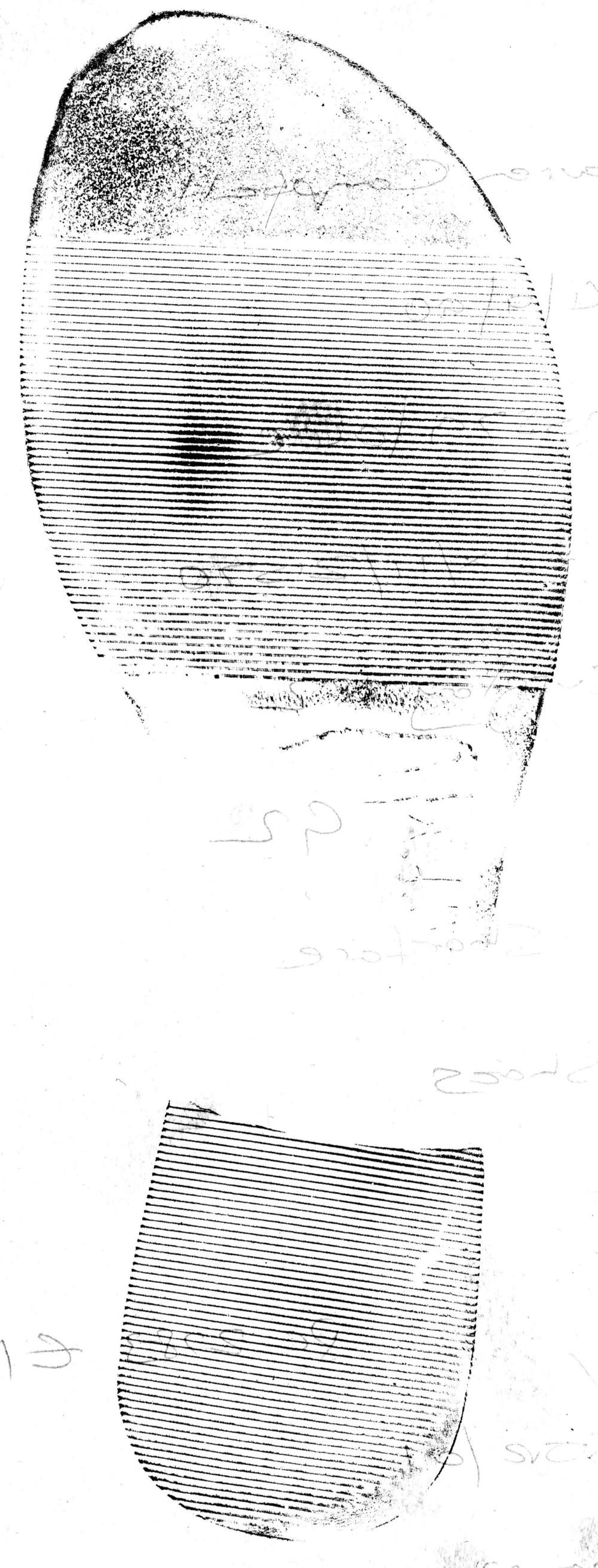}}
\subfigure[$H_{(1000 \times 2620)}$]{\label{fig:h_5}\includegraphics[width=32mm]{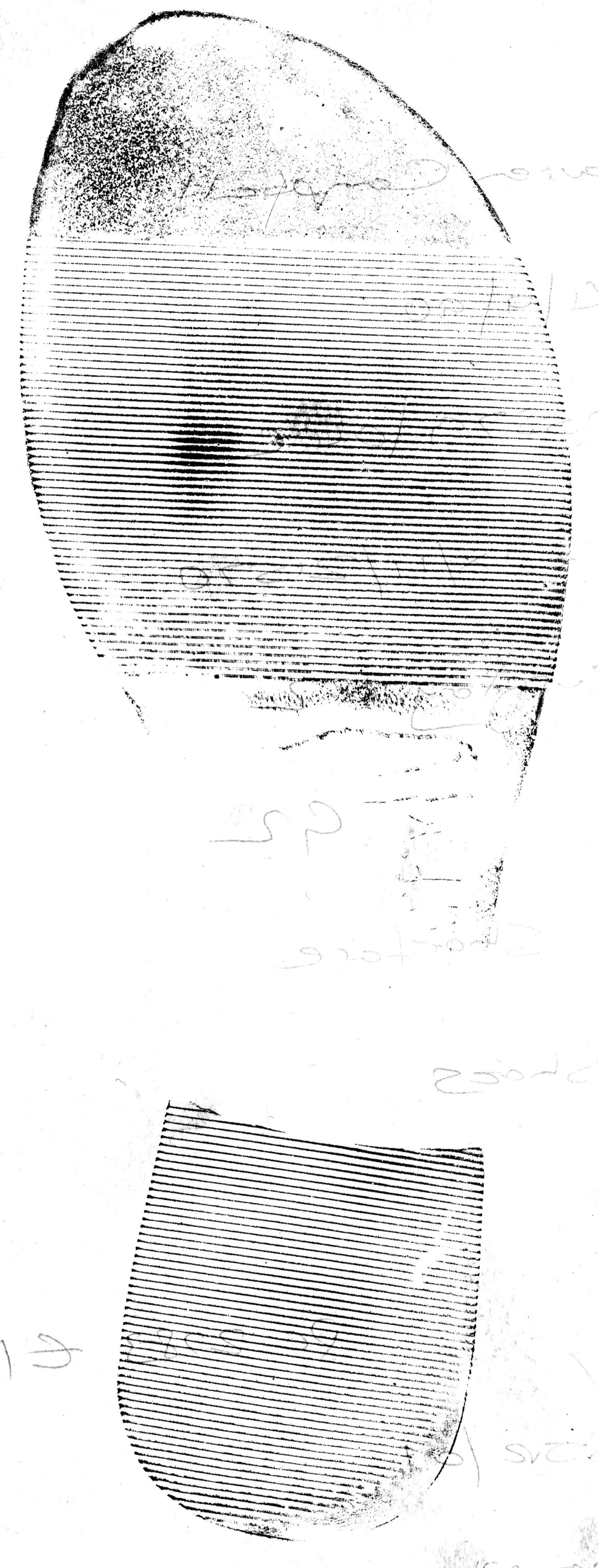}}

\caption{Interpolation samples with fixed aspect ratio and varying sizes (zoom in to see the original pattern)}
\label{fig:interpolation_samples_3}
\end{figure}

Although an ideal interpolation technique is expected not to alter any pattern within the image or introduce any artefact, most of the interpolation techniques usually alter some image features and also introduce artefacts when interpolated to reduce image resolution~\cite{thevenaz2000interpolation, tsao2003interpolation}. 
Figures~\ref{fig:interpolation_samples_1},~\ref{fig:interpolation_samples_2}, and~\ref{fig:interpolation_samples_3} shows how the interpolation techniques discussed above can affect the features of a footwear impression image at different resolutions\footnote{Please zoom in to see how the interpolation pattern gradually resembles the original image (Figure~\ref{fig:original_no_interpolation}) with increasing resolution. Zooming is required as the images embedded in this paper go through arbitrary interpolation applied by your browser or PDF reader.}. It is apparent from the undersampled (Figure~\ref{fig:interpolation_samples_1}) images that using different interpolation techniques produce slightly different images. Although these discrepancies are aesthetically undesirable and can hamper the performance of the model, we leverage such differences as an effective image augmentation technique as described in Section~\ref{sec:Experimental Setup}. 

As we can see, all three interpolated images closely resemble the original (Figure~\ref{fig:original_no_interpolation}) at a higher resolution (Figure~\ref{fig:interpolation_samples_3}) and at the same time their differences reduces. Comparing between the lowest resolution images (Figure~\ref{fig:interpolation_samples_1}), it is apparent that Nearest Neighbour (N) produces the most different looking downsampled image. Additionally, all the lower resolution impression images produce~\emph{descriptor}~\emph{D01: Bars} which are angled whereas the original (Figure~\ref{fig:original_no_interpolation}) and higher resolution (zoom in for Figure~\ref{fig:interpolation_samples_3}) impressions have~\emph{descriptor}~\emph{D01: Bars} which are straight lines with an angle of $0^{\circ}$. A study by~\cite{koziarski2018impact} found that even a mildest quality loss of input images can greatly hamper the performance of a deep learning model. \citet{glorot2010understanding} too found neural networks to be susceptible to image noise. 

\section{Experimental Setup}
\label{sec:Experimental Setup}
In our experiments, we use \textit{ResNet-50}, a popular 50 layer CNN architecture with residual connections~\cite{he2016deep}, pre-trained on the ImageNet dataset~\cite{deng2009imagenet} with a custom head initialised using the Glorot/Xavier initialisation~\cite{glorot2010understanding} and optional, learnable preprocessing layer (see below). The head consists of an adaptive pooling layer, followed by two \textit{BatchNorm} $\rightarrow$ \textit{Dropout} $\rightarrow$ \textit{Linear/Dense} blocks with \textit{ReLU} non-linearity in between. The number of units in the non-output linear layer was set to $512$, while the output layer has a total of 17 neurons with \textit{sigmoid} activation functions, one per each \emph{descriptor} type. The models are trained using \textit{AdamW}~\cite{loshchilov2017decoupled}, a stochastic gradient descent based backpropagation algorithm in two phases:
\begin{enumerate}
    \item \textbf{Initial training}, where all the \textit{ResNet-50} body layers are frozen and only the 2-layer head as well as the optional preprocessing layer are trained with the learning rate of $1e-3$ and weight decay of $0.1$.
    \item\textbf{Fine-tuning}, where the whole network is trained using discriminative learning rates~\cite{howard2018universal} of between $1e-6$ and $1e-4$ and weight decay of $0.1$.
\end{enumerate}

\begin{figure}[H]
\centering
\includegraphics[width=0.75\textwidth]{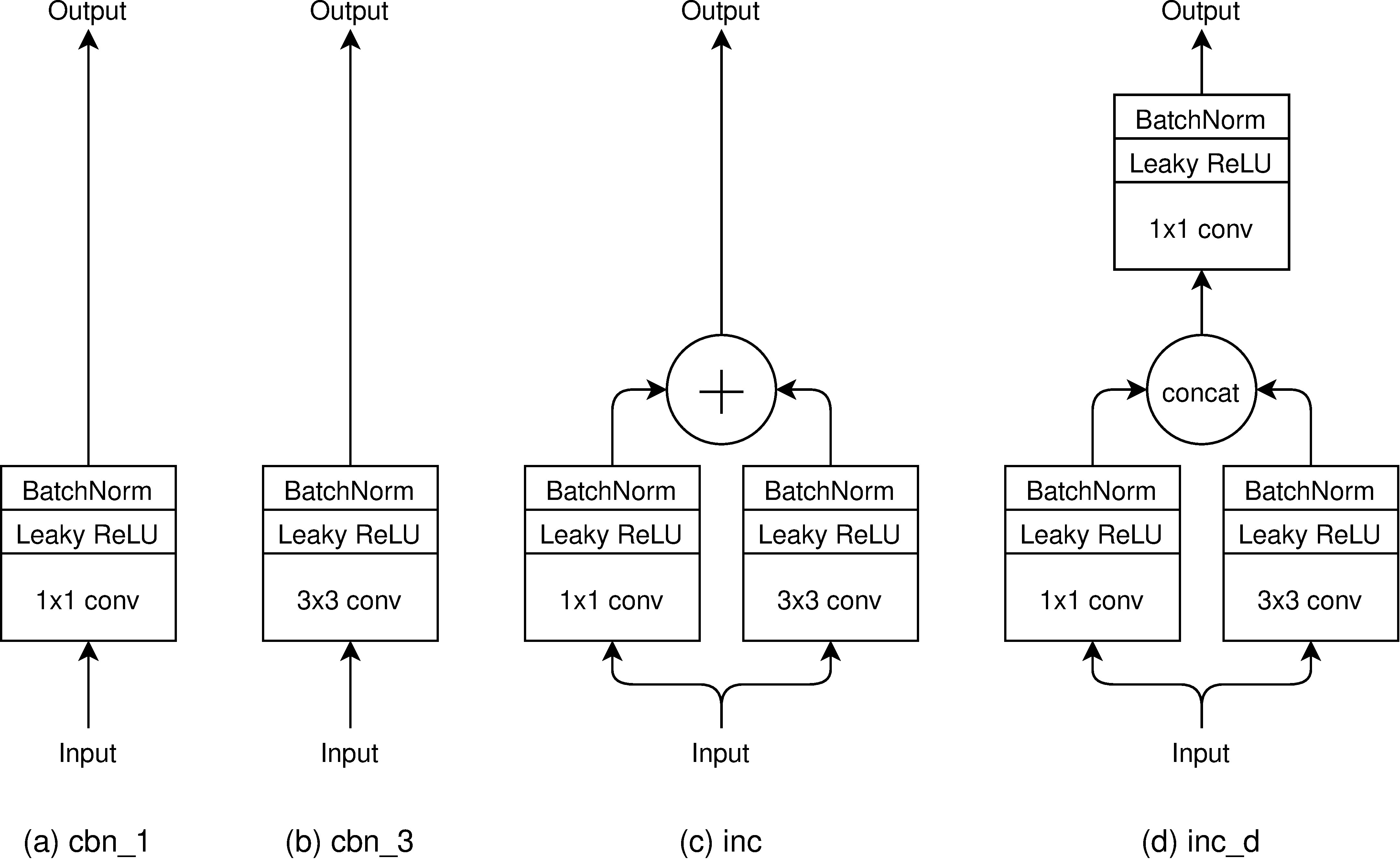}
\caption{Learnable preprocessing layers: (a)~\textbf{cbn\_1}: 1x1 Conv and BatchNorm, (b)~\textbf{cbn\_3}: 3x3 Conv and BatchNorm, (c)~\textbf{inc}: inception-like transformation, (d)~\textbf{inc\_d}: dense inception-like transformation}
\label{fig:transforms}
\end{figure}

We have experimented with various combinations of the following:
\begin{enumerate}
    \item \textbf{Loss function}: In addition to the default Binary Cross Entropy (BCE) loss, which in our experiments was always used in a cost-sensitive setting via class weighting (i.e. with the cost of misclassification being inversely proportional to class frequency in the training dataset), we have also used the Soft-F1 loss in an attempt to maximise both precision and recall directly within the model training process. The Soft-F1 loss is a simple generalisation of the F1 score obtained by replacing the number of True Positives (TP), False Positives (FP) and False Negatives (FN) with their probabilistic counterparts~\cite{eban2017scalable}:
    

    \begin{align*} 
    TP &= \sum_i y_i    \hat{y}_i \\
    FP &= \sum_i (1-y_i)\hat{y}_i \\
    FN &= \sum_i y_i (1-\hat{y}_i)
    \end{align*}
    where $y_i \in \{0,1\}$ is the label for the $i^{th}$ data instance and $\hat{y_i} \in \interval[]{0}{1}$ is the model prediction. 
    
    \item \textbf{Channel configuration}: All the original input images are greyscale (single-channel), yet the pre-trained model expects RGB/colour inputs (three-channels). The simplest and most popular approach to address this discrepancy is to collate three identical copies of the greyscale input. Since this approach seems wasteful, we have instead opted for various compositions of the three-channel input obtained via applying \textit{different} interpolation techniques (see Section~\ref{sec:Image Interpolation Techniques}) to the high resolution input image -- these are specified in Table~\ref{tab:interpolation_combinations}.
    \item \textbf{Preprocessing layer}: For the same reasons as described above, we have included a number of learnable preprocessing layers in our network. The rationale here was that the distribution of greyscale images, particularly when collating three different interpolated versions of each image into a single three-channel input, is different from the distribution of natural RGB images from the ImageNet dataset. The preprocessing layers we have used have been shown in Figure~\ref{fig:transforms}. 
\end{enumerate}

\begin{table}[H]
\footnotesize
\begin{center}
\begin{tabular}{|c|c|c|c|}
\hline
\textbf{channels} & \textbf{R}          & \textbf{G}        & \textbf{B}        \\ \hline
B-B-B             & Bilinear            & Bilinear          & Bilinear          \\ \hline
B-H-N             & Bilinear            & Hamming           & Nearest Neighbour \\ \hline
B-N-H             & Bilinear            & Nearest Neighbour & Hamming           \\ \hline
H-B-N             & Hamming             & Bilinear          & Nearest Neighbour \\ \hline
H-H-H             & Hamming             & Hamming           & Hamming           \\ \hline
H-N-B             & Hamming             & Nearest Neighbour & Bilinear          \\ \hline
N-B-H             & Nearest Neighbour   & Bilinear          & Hamming           \\ \hline
N-H-B             & Nearest Neighbour   & Hamming           & Bilinear          \\ \hline
N-N-N             & Nearest Neighbour   & Nearest Neighbour & Nearest Neighbour \\ \hline
\end{tabular}
\caption{Compositions of input channels via different combinations of interpolation techniques}
\label{tab:interpolation_combinations}
\end{center}
\end{table}

\begin{figure}[H]
\centering
\subfigure[Training set]{\includegraphics[width=0.50\textwidth]{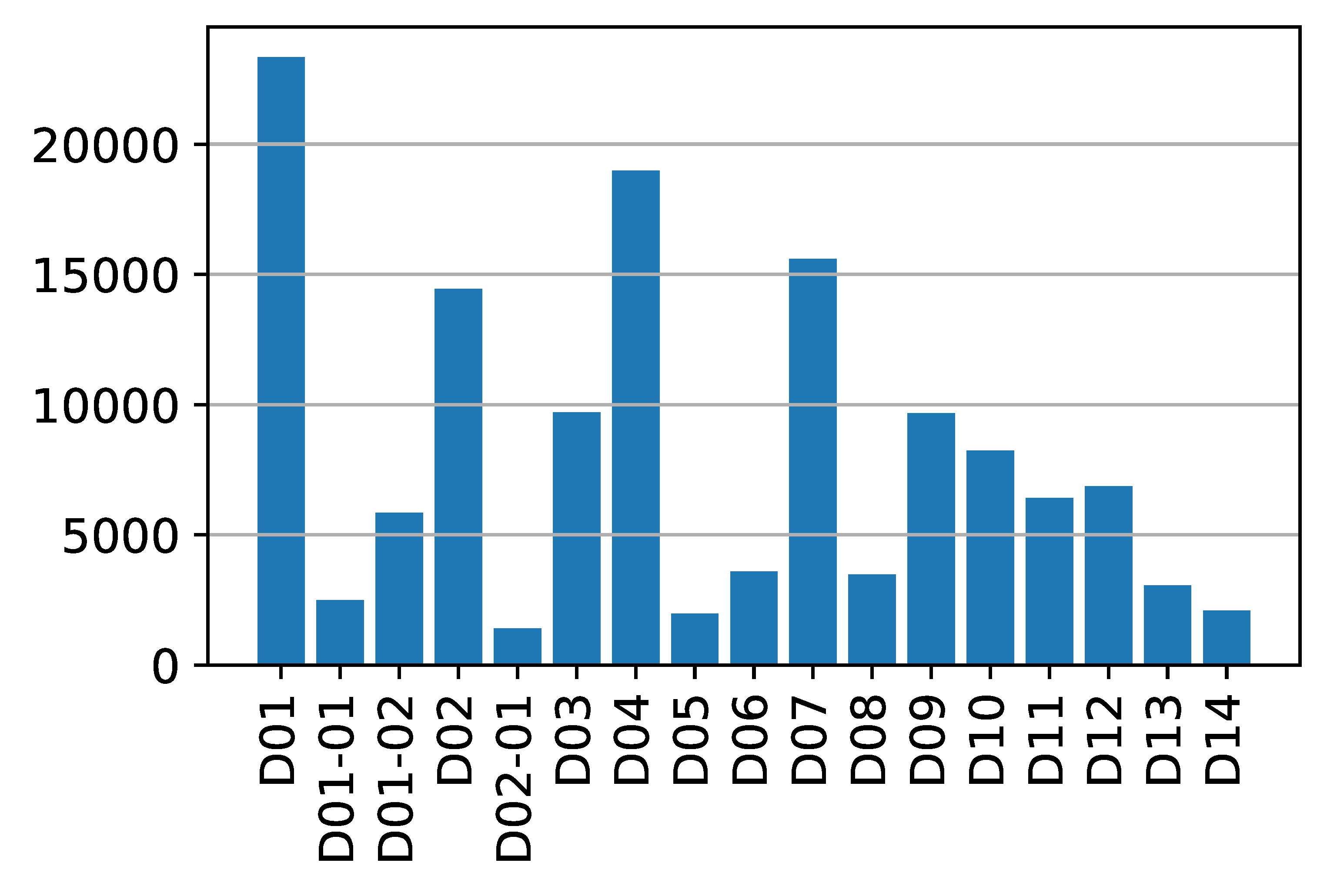}}
\subfigure[Validation set]{\includegraphics[width=0.48\textwidth]{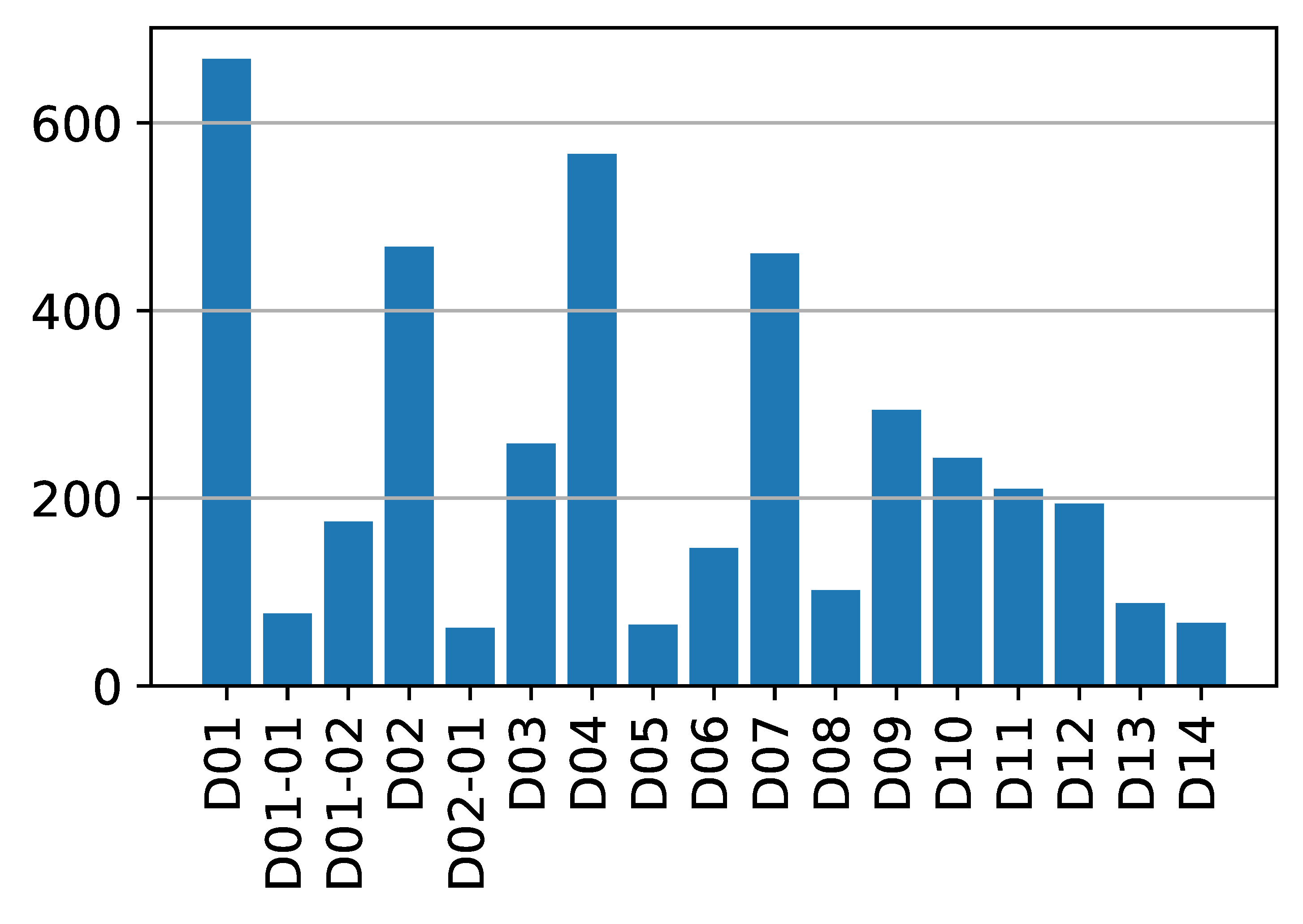}}
\caption{Class distribution}
\label{fig:class_hist}
\end{figure}

\begin{figure}[H]
\centering
\includegraphics[width=0.9\textwidth]{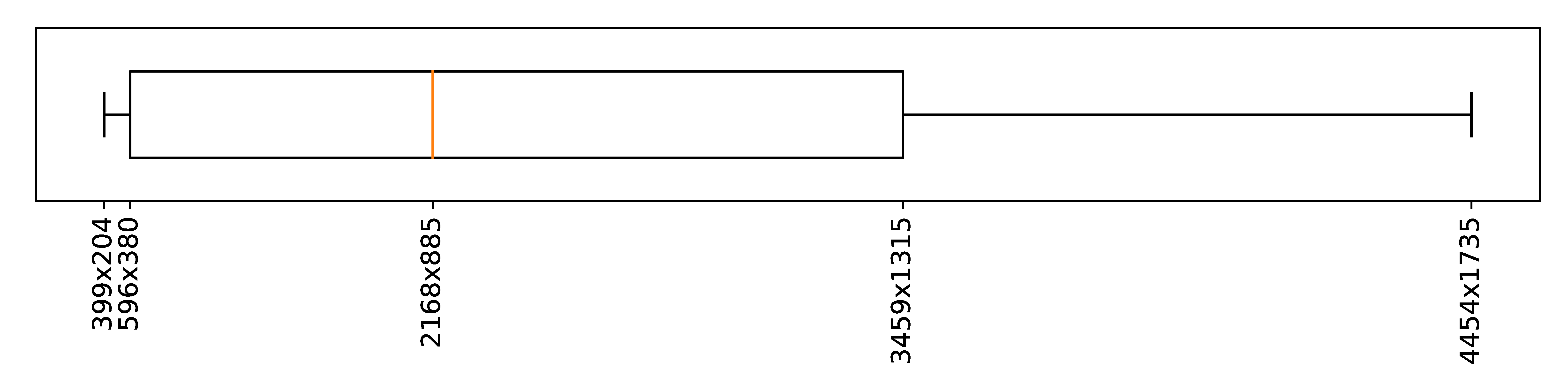}
\caption{Original input image resolution}
\label{fig:res_boxplot}
\end{figure}

The training dataset consisted of 33,757 greyscale images retrieved from the NFRC with the class distribution as shown in Figure~\ref{fig:class_hist}a. As it can be seen, the classes are dominated by \textit{D01}, \textit{D02}, \textit{D04} and \textit{D07}, with \textit{D01-01}, \textit{D02-01}, \textit{D05} and \textit{D14} being the least frequent. The validation set consisted of 1,000 images retrieved from the same database with the class distribution as shown in Figure~\ref{fig:class_hist}b. The image resolution ranged from $180 \times 60$ to $15,000 \times 7,000$ and has been depicted in Figure~\ref{fig:res_boxplot}, where the whiskers represent $Q_{05}$ and $Q_{95}$, and outliers have been omitted for presentation clarity.

\section{Results}\label{sec:Results}

Table~\ref{tab:my-table} contains the aggregated results of the total of 180 experiments run across 90 different combinations of hyper-parameters as specified in Table~\ref{tab:interpolation_combinations} and Figure~\ref{fig:transforms}. Each experiment has been repeated twice with random initialisation, and the average of these two runs was used to construct Table~\ref{tab:my-table}. In each of the experiments, we have trained a custom head on top of a fixed/frozen ImageNet pre-trained ResNet-50 for 10 epochs, followed by 40 epochs of finetuning of the whole network. We have opted for the Area Under the Precision-Recall Curve (\textit{PRAUC}) as the performance metric in order to decouple the results from class-specific thresholds. 
\textit{PRAUC} is a better measurement of performance of a binary classifier than the \textit{AUC} of the \textit{ROC} (receiver operating characteristic) curve~\cite{saito2015precision,wahid2019predict} since \textit{ROC} is very sensitive to class imbalance and in our case class labels are heavily imbalanced. 

\begin{table}[]
\footnotesize
\begin{center}
\begin{adjustbox}{width=0.81\textwidth}
\begin{tabular}{|c|c||c|c|c||c|c|c|}
\hline
\multirow{3}{*}{\textbf{channels}} & \multirow{3}{*}{\textbf{preprocessing}} & \multicolumn{6}{c|}{\textbf{PRAUC}} \\ \cline{3-8} & & \multicolumn{3}{c||}{\textbf{BCE LOSS}} & \multicolumn{3}{c|}{\textbf{F1 LOSS}} \\ \cline{3-8} & & \textbf{352x144} & \textbf{224x224} & \textbf{$\Delta$} & \textbf{352x144}  & \textbf{224x224} & \textbf{$\Delta$}  \\ \hline\hline
N-N-N & cbn\_1  & 0.7200 & 0.6969 & 0.0231 & 0.6927 & 0.6712 & 0.0215 \\ \hline
N-N-N & cbn\_3  & 0.7171 & 0.6968 & 0.0203 & 0.6839 & 0.6741 & 0.0098 \\ \hline
N-N-N & inc     & 0.7205 & 0.6990 & 0.0215 & 0.6861 & 0.6706 & 0.0155 \\ \hline
N-N-N & inc\_d  & 0.7215 & 0.6945 & 0.0270 & 0.6938 & 0.6712 & 0.0226 \\ \hline
N-N-N & no\_tfm & 0.7135 & 0.6940 & 0.0195 & 0.6902 & 0.6728 & 0.0174 \\ \hline
H-H-H & cbn\_1  & 0.7113 & 0.6958 & 0.0155 & 0.6876 & 0.6694 & 0.0182\\ \hline
H-H-H & cbn\_3  & 0.7172 & 0.6939 & 0.0233 & 0.6852 & \textbf{0.6778} & \textbf{\underline{0.0074}} \\ \hline
H-H-H & inc     & 0.7133 & 0.6931 & 0.0202 & 0.6878 & 0.6721 & 0.0157 \\ \hline
H-H-H & inc\_d  & \textbf{0.7259} & \textbf{0.7044} & 0.0215 & 0.6923 & 0.6753 & 0.0170 \\ \hline
H-H-H & no\_tfm & 0.7169 & 0.6995 & 0.0174 & 0.6874 & 0.6732 & 0.0142 \\ \hline
B-B-B & cbn\_1  & 0.7177 & 0.6956 & 0.0221 & 0.6872 & 0.6733 & 0.0139 \\ \hline
B-B-B & cbn\_3  & 0.7165 & 0.6974 & 0.0191 & 0.6892 & 0.6710 & 0.0182 \\ \hline
B-B-B & inc     & 0.7176 & 0.6976 & 0.0200 & 0.6914 & 0.6705 & 0.0209\\ \hline
B-B-B & inc\_d  & 0.7198 & 0.699  & 0.0208 & 0.6960 & 0.6679 & \textbf{0.0281} \\ \hline
B-B-B & no\_tfm & \textbf{\underline{0.7061}} & 0.6913 & 0.0148 & 0.6895 & 0.6649 & 0.0246 \\ \hline
N-B-H & cbn\_1  & 0.7198 & 0.6981 & 0.0217 & 0.6879 & 0.6702 & 0.0177 \\ \hline
N-B-H & cbn\_3  & 0.7118 & 0.6958 & 0.0160 & 0.6879 & 0.6690 & 0.0189 \\ \hline
N-B-H & inc     & 0.7127 & 0.6957 & 0.0170 & 0.6851 & 0.6703 & 0.0148 \\ \hline
N-B-H & inc\_d  & 0.7204 & 0.6931 & 0.0273 & 0.6899 & 0.6725 & 0.0174 \\ \hline
N-B-H & no\_tfm & 0.7158 & 0.6972 & 0.0186 & 0.6878 & 0.6705 & 0.0173 \\ \hline
N-H-B & cbn\_1  & 0.7118 & 0.6919 & 0.0199 & 0.6894 & 0.6665 & 0.0229 \\ \hline
N-H-B & cbn\_3  & 0.7201 & 0.7005 & 0.0196 & 0.6890 & 0.6732 & 0.0158 \\ \hline
N-H-B & inc     & 0.7161 & 0.6973 & 0.0188 & 0.6943 & 0.6701 & 0.0242 \\ \hline
N-H-B & inc\_d  & 0.7189 & 0.6988 & 0.0201 & 0.6965 & 0.6750 & 0.0215 \\ \hline
N-H-B & no\_tfm & 0.7144 & 0.6971 & 0.0173 & 0.6896 & 0.6675 & 0.0221 \\ \hline
B-H-N & cbn\_1  & 0.7167 & 0.6953 & 0.0214 & 0.6893 & 0.6698 & 0.0195 \\ \hline
B-H-N & cbn\_3  & 0.7200 & 0.7007 & 0.0193 & 0.6895 & 0.6734 & 0.0161 \\ \hline
B-H-N & inc     & 0.7206 & \textbf{\underline{0.6907}} & \textbf{0.0299} & 0.6865 & 0.6724 & 0.0141 \\ \hline
B-H-N & inc\_d  & 0.7156 & 0.6976 & 0.0180 & 0.6896 & 0.6697 & 0.0199 \\ \hline
B-H-N & no\_tfm & 0.7092 & 0.6972 & \textbf{\underline{0.0120}} & 0.6914 & 0.6720 & 0.0194 \\ \hline
B-N-H & cbn\_1  & 0.7169 & 0.7011 & 0.0158 & 0.6863 & 0.6709 & 0.0154 \\ \hline
B-N-H & cbn\_3  & 0.7187 & 0.6991 & 0.0196 & 0.6866 & 0.6670 & 0.0196 \\ \hline
B-N-H & inc     & 0.7177 & 0.6977 & 0.0200 & 0.6854 & 0.6761 & 0.0093 \\ \hline
B-N-H & inc\_d  & 0.7232 & 0.7016 & 0.0216 & 0.6874 & \textbf{\underline{0.6642}} & 0.0232 \\ \hline
B-N-H & no\_tfm & 0.7149 & 0.6968 & 0.0181 & 0.6891 & 0.6711 & 0.0180 \\ \hline
H-N-B & cbn\_1  & 0.7178 & 0.6982 & 0.0196 & 0.6875 & 0.6704 & 0.0171 \\ \hline
H-N-B & cbn\_3  & 0.7124 & 0.6963 & 0.0161 & 0.6875 & 0.6667 & 0.0208 \\ \hline
H-N-B & inc     & 0.7134 & 0.6985 & 0.0149 & \textbf{\underline{0.6849}} & 0.6728 & 0.0121 \\ \hline
H-N-B & inc\_d  & 0.7176 & 0.6984 & 0.0192 & 0.6896 & 0.6693 & 0.0203 \\ \hline
H-N-B & no\_tfm & 0.7107 & 0.6987 & 0.0120 & 0.6897 & 0.6678 & 0.0219 \\ \hline
H-B-N & cbn\_1  & 0.7224 & 0.6929 & 0.0295 & 0.6889 & 0.6698 & 0.0191 \\ \hline
H-B-N & cbn\_3  & 0.7171 & 0.6939 & 0.0232 & 0.6850 & 0.6732 & 0.0118 \\ \hline
H-B-N & inc     & 0.7146 & 0.7007 & 0.0139 & 0.6861 & 0.6685 & 0.0176 \\ \hline
H-B-N & inc\_d  & 0.7242 & 0.6975 & 0.0267 & \textbf{0.6966} & 0.6751 & 0.0215 \\ \hline
H-B-N & no\_tfm & 0.7118 & 0.6915 & 0.0203 & 0.6865 & 0.6706 & 0.0159 \\ \hline\hline
\multicolumn{2}{|c||}{$\mu$}  & 0.7167 & 0.6969 & 0.0199 & 0.6889 & 0.6709 & 0.0180 \\ \hline
\end{tabular}
\end{adjustbox}
\end{center}
\caption{Performance based on the Area Under the Precision-Recall Curve (\textit{PRAUC}). `$\Delta$' denotes difference between the two resolutions. \textit{Max} for each column in \textbf{bold}, \textit{min} in \textbf{\underline{underline}}.
}
\label{tab:my-table}
\end{table}

The first thing to notice is that according to the results, preserving the aspect ratio of the input images (i.e. $352 \times 144$ resolution) \textbf{always} leads to better performance than when using squished images (i.e. $224 \times 224$ resolution, see the `$\Delta$' column in Table~\ref{tab:my-table}). This holds regardless of the type of resampling, preprocessing and loss function used. It appears that preserving the aspect ratio of the original images matters much more than higher horizontal resolution (i.e. 144 vs 224 pixels) and is the best way of spending a fixed computational budget (the number of input pixels in both cases is approximately equal ($224 \times 224=50,176$, $352 \times 144=50,688$). Although this performance difference might be partially attributed to the nature of our inputs (shoe impressions are long and thin, so squishing can introduce significant distortions), the same can be said about many other objects like people or vehicles. It is hence somewhat surprising that $224 \times 224$ is the default transfer learning setting for popular deep learning frameworks~\cite{chollet2015keras,paszke2017automatic,abadi2016tensorflow}. 

The average difference in \textit{PRAUC} between the two resolutions (everything else being equal) is 0.0189 (2.8\%), while the maximum difference over all 180 runs reaches 0.0415 (6.5\%). To put these numbers in context, the average difference between any two randomly initialised runs of each experiment is 0.0051, while the maximum difference is 0.0206. Thus the observed effect is unlikely to be a random fluctuation. For this reason, we limit further analysis to the results for the $352 \times 144$ (and higher) resolutions, preserving the input aspect ratio.

A similar observation can be made when considering the loss function. \textit{BCE} \textbf{consistently} outperforms the \textit{F1 LOSS}, with the average \textit{PRAUC} difference of 0.0269 and the maximum difference of 0.0374. Despite attractive theoretical properties of the \textit{F1 LOSS} as discussed in Section~\ref{sec:Experimental Setup}, the \textit{BCE} loss proved to be a much better choice in practice. For this reason we are not considering the \textit{F1 LOSS} in the subsequent analysis.


\begin{table}[h]
\scriptsize
\begin{minipage}{.02\linewidth}\quad\end{minipage} 
\begin{minipage}{.45\linewidth}
\centering
    \begin{tabular}{|c|c|c|}
    \hline
    \textbf{rank} & \textbf{channels} & \textbf{PRAUC (↓)}  \\ \hline
    1             & N-N-N             & 0.7185          \\ \hline
    2             & B-N-H             & 0.7183          \\ \hline
    3             & H-B-N             & 0.7180          \\ \hline
    4             & H-H-H             & 0.7169          \\ \hline
    5             & B-H-N             & 0.7164          \\ \hline
    6             & N-H-B             & 0.7163          \\ \hline
    7             & N-B-H             & 0.7161          \\ \hline
    8             & B-B-B             & 0.7155          \\ \hline
    9             & H-N-B             & 0.7144          \\ \hline\hline
    \multicolumn{2}{|c|}{$\max(\Delta)$} & 0.0041       \\ \hline
    \end{tabular}
\caption{Performance for $352 \times 144$ input resolution, \textit{BCE} loss, and various input channel configurations, averaged over preprocessing methods, sorted by \textit{PRAUC}.\label{tab:channels_PRAUC}}
\end{minipage}
\begin{minipage}{.06\linewidth}\quad\end{minipage} 
\begin{minipage}{.45\linewidth}
\centering
    \begin{tabular}{|c|c|c|}
    \hline
    \textbf{rank} & \textbf{preprocessing} & \textbf{PRAUC (↓)} \\ \hline
    1             & inc\_d                 & 0.7208           \\ \hline
    2             & cbn\_1                 & 0.7171           \\ \hline
    3             & cbn\_3                 & 0.7168           \\ \hline
    4             & inc                    & 0.7163           \\ \hline
    5             & no\_tfm                & 0.7126           \\ \hline\hline
    \multicolumn{2}{|c|}{$\max(\Delta)$}      & 0.0082           \\ \hline
    \end{tabular}
\caption{Performance for $352 \times 144$ input resolution, \textit{BCE} loss, and various preprocessing layers, averaged over input channel configurations, sorted by \textit{PRAUC}.\label{tab:transform_PRAUC}}
\end{minipage}
\begin{minipage}{.02\linewidth}\quad\end{minipage} 
\end{table}

The average difference in \textit{PRAUC} among various combinations of input channels (Table~\ref{tab:channels_PRAUC}) is much less pronounced. The approach of creating an input to the network by `sandwiching' the outputs of three different interpolation methods rather than simply creating three identical channels (i.e. \textit{B-N-H} vs \textit{N-N-N}), doesn't seem to affect the \textit{PRAUC} much, with the difference between the best and worst performing approach being 0.0041. Nevertheless, it's worth noting that the \textit{B-B-B} approach, which is the default in the existing deep learning frameworks~\cite{chollet2015keras,paszke2017automatic,abadi2016tensorflow}, is one of the worst performing. This seems to confirm the intuition that the blurring effect characteristic for bilinear interpolation, tends to make discrimination between different types of descriptors more challenging. Another observation is that the influence of input channel ordering on \textit{PRAUC} can be almost as big as the difference between the best and worst performing approach (the difference between \textit{B-N-H} and \textit{H-N-B} is 0.0039). This is somewhat surprising as in theory, the learnable preprocessing layer should be able to `swap' the input channel order if needed. However, in the context of the average difference between any two randomly initialised runs of each experiment, which as mentioned earlier was 0.0051, the results given in Table~\ref{tab:channels_PRAUC} need to be declared inconclusive. 

The influence of the learnable preprocessing layer on \textit{PRAUC} is more substantial. As it can be seen in Table~\ref{tab:transform_PRAUC}, the difference between the dense inception-like transformation (\textit{inc\_d}) and no transformation at all (\textit{no\_tfm}) reaches 0.0082. Since \textit{no\_tfm} is the worst performing approach in our experiments, we conclude that using some kind of learnable preprocessing is beneficial.

\begin{table}[!ht]
\scriptsize
\begin{center}
\begin{tabular}{c|c|c|c|c|c|cc}
\cline{2-6}
\multicolumn{1}{l|}{} & \multicolumn{5}{c|}{\textbf{preprocessing}} \\ \hline
\multicolumn{1}{|c||}{\textbf{channels}} & \textbf{cbn\_1} & \textbf{cbn\_3} & \textbf{inc} & \textbf{inc\_d} & \textbf{no\_tfm} & \multicolumn{1}{c||}{$\mu$}  & \multicolumn{1}{c|}{$\max(\Delta)$} \\ \hline\hline
\multicolumn{1}{|c||}{\textbf{B-B-B}}    & 0.7177               & 0.7165               & 0.7176          & \underline{0.7198}                 & 0.7061           & \multicolumn{1}{c||}{0.7155} & \multicolumn{1}{c|}{0.0137}         \\ \hline
\multicolumn{1}{|c||}{\textbf{B-H-N}}    & 0.7167               & 0.7200               & \textbf{\underline{0.7206}}          & 0.7156                 & 0.7092           & \multicolumn{1}{c||}{0.7164} & \multicolumn{1}{c|}{0.0114}         \\ \hline
\multicolumn{1}{|c||}{\textbf{B-N-H}}    & 0.7169               & 0.7187               & 0.7177          & \underline{0.7232}                 & 0.7149           & \multicolumn{1}{c||}{0.7183} & \multicolumn{1}{c|}{0.0083}         \\ \hline
\multicolumn{1}{|c||}{\textbf{H-B-N}}    & \textbf{0.7224}               & 0.7171               & 0.7146          & \underline{0.7242}                 & 0.7118           & \multicolumn{1}{c||}{0.7180} & \multicolumn{1}{c|}{0.0124}         \\ \hline
\multicolumn{1}{|c||}{\textbf{H-H-H}}    & 0.7113               & 0.7172               & 0.7133          & \textbf{\underline{0.7259}}                 & \textbf{0.7169}           & \multicolumn{1}{c||}{0.7169} & \multicolumn{1}{c|}{0.0146}         \\ \hline
\multicolumn{1}{|c||}{\textbf{H-N-B}}    & \underline{0.7178}               & 0.7124               & 0.7134          & 0.7176                 & 0.7107           & \multicolumn{1}{c||}{0.7144} & \multicolumn{1}{c|}{0.0071}         \\ \hline
\multicolumn{1}{|c||}{\textbf{N-B-H}}    & 0.7198               & 0.7118               & 0.7127          & \underline{0.7204}                 & 0.7158           & \multicolumn{1}{c||}{0.7161} & \multicolumn{1}{c|}{0.0086}         \\ \hline
\multicolumn{1}{|c||}{\textbf{N-H-B}}    & 0.7118               & \textbf{\underline{0.7201}}               & 0.7161          & 0.7189                 & 0.7144           & \multicolumn{1}{c||}{0.7163} & \multicolumn{1}{c|}{0.0083}         \\ \hline
\multicolumn{1}{|c||}{\textbf{N-N-N}}    & 0.7200               & 0.7171               & 0.7205          & \underline{0.7215}                 & 0.7135           & \multicolumn{1}{c||}{\textbf{0.7185}} & \multicolumn{1}{c|}{0.0080}         \\ \hline\hline
\multicolumn{1}{|c||}{$\mu$}             & 0.7171               & 0.7168               & 0.7163          & \underline{0.7208}                 & 0.7126           & \multicolumn{1}{c||}{0.7167} & \multicolumn{1}{l}{}                \\ \cline{1-7}
\multicolumn{1}{|c||}{$\max(\Delta)$}    & 0.0111               & 0.0083               & 0.0079          & 0.0103                 & 0.0076           & \multicolumn{1}{l}{}        & \multicolumn{1}{l}{}                \\ \cline{1-6}
\end{tabular}
\caption{\textit{PRAUC} for $352 \times 144$ input resolution, BCE loss, and various combinations of pre-first layer transformations and input channel configurations. The highest score in each column in \textbf{bold}. The highest \textit{PRAUC} in each row in \underline{underline}.}
\label{tab:transform_channel_PRAUC}
\end{center}
\end{table}

Table~\ref{tab:transform_channel_PRAUC} presents the breakdown of the results by preprocessing layers and input channel configurations. As it can be seen, \textit{inc\_d} gives the highest PRAUC on average (0.7208), is the best preprocessing method for 6 out of 9 input channel configurations, and second best for additional 2. It is harder to identify the best performing channel configuration as none of them seems to be dominating across different preprocessing layers. However, looking at preprocessing and channel configuration jointly, \textit{inc\_d} with \textit{H-H-H} gives the highest \textit{PRAUC} of 0.7259, which is 0.0198 more than the worst performing combination (\textit{no\_tfm} with \textit{B-B-B}) and 0.0092 more than the average across all the entries in Table~\ref{tab:transform_channel_PRAUC}.

\begin{table}[!ht]
\scriptsize
\begin{center}
\begin{tabular}{c|c|c|c|c|cc}
\cline{2-5}
\multicolumn{1}{l|}{} & \multicolumn{4}{c|}{\textbf{preprocessing}} \\ \hline
\multicolumn{1}{|c||}{\textbf{channels}} & \textbf{cbn\_1} & \textbf{cbn\_3} & \textbf{inc} & \textbf{inc\_d} & \multicolumn{1}{c||}{$\mu$}  & \multicolumn{1}{c|}{$\max(\Delta)$} \\ \hline\hline

\multicolumn{1}{|c||}{\textbf{B-B-B}}    & \underline{\textbf{0.7450}}              & 0.7407               & 0.7363          & 0.7369                 & \multicolumn{1}{c||}{0.7397} & \multicolumn{1}{c|}{0.0086}         \\ \hline
\multicolumn{1}{|c||}{\textbf{B-H-N}}    & 0.7394               & 0.7334               & 0.7371          & \underline{0.7400}                 & \multicolumn{1}{c||}{0.7375} & \multicolumn{1}{c|}{0.0067}         \\ \hline
\multicolumn{1}{|c||}{\textbf{B-N-H}}    & \underline{0.7399}               & 0.7365               & 0.7375          & \underline{0.7399}                 & \multicolumn{1}{c||}{0.7384} & \multicolumn{1}{c|}{0.0035}         \\ \hline
\multicolumn{1}{|c||}{\textbf{H-B-N}}    & 0.7407               & 0.7409               & 0.7377          & \underline{\textbf{0.7477}}                 & \multicolumn{1}{c||}{0.7410} & \multicolumn{1}{c|}{0.0070}         \\ \hline
\multicolumn{1}{|c||}{\textbf{H-H-H}}    & \underline{\textbf{0.7450}}               & 0.7404               & 0.7361          & 0.7441                 & \multicolumn{1}{c||}{0.7414} & \multicolumn{1}{c|}{0.0089}         \\ \hline
\multicolumn{1}{|c||}{\textbf{H-N-B}}    & 0.7405               & \underline{0.7425}               & 0.7363          & 0.7411                 & \multicolumn{1}{c||}{0.7401} & \multicolumn{1}{c|}{0.0062}         \\ \hline
\multicolumn{1}{|c||}{\textbf{N-B-H}}    & 0.7383               & 0.7311               & \underline{\textbf{0.7436}}          & 0.7432                 & \multicolumn{1}{c||}{0.7390} & \multicolumn{1}{c|}{0.0125}         \\ \hline
\multicolumn{1}{|c||}{\textbf{N-H-B}}    & 0.7367               & \textbf{0.7429}               & 0.7415          & \underline{0.7451}                 & \multicolumn{1}{c||}{\textbf{0.7415}} & \multicolumn{1}{c|}{0.0084}         \\ \hline
\multicolumn{1}{|c||}{\textbf{N-N-N}}    & 0.7440               & 0.7303               & 0.7399          & \underline{0.7459}                 & \multicolumn{1}{c||}{0.7400} & \multicolumn{1}{c|}{0.0156}         \\ \hline\hline
\multicolumn{1}{|c||}{$\mu$}             & 0.7410               & 0.7376               & 0.7385          & \underline{0.7423}                 & \multicolumn{1}{c||}{0.7399} & \multicolumn{1}{l}{}                \\ \cline{1-6}
\multicolumn{1}{|c||}{$\max(\Delta)$}    & 0.0083               & 0.0126               & 0.0075          & 0.0090                 & \multicolumn{1}{l}{}        & \multicolumn{1}{l}{}                \\ \cline{1-5}
\end{tabular}
\caption{\textit{PRAUC} for $464 \times 192$ input resolution, BCE loss, and various combinations of preprocessing layers and input channel configurations. The highest score in each column in \textbf{bold}. The highest \textit{PRAUC} in each row in \underline{underline}.}
\label{tab:transform_channel_PRAUC_hires}
\end{center}
\end{table}

In Table~\ref{tab:transform_channel_PRAUC_hires} we report the results of a similar experiment, this time with the resolution of the input images increased to $464 \times 192$. This lead to a significant increase of the \textit{PRAUC} across all tested combinations of preprocessing layers and input channel configurations, with the minimum, average and maximum difference of 0.0132, 0.0221 and 0.0337 respectively. As before, \textit{inc\_d} is the dominating preprocessing method, while none of the input channel configurations seems to be a clear winner. Note, that in Table~\ref{tab:transform_channel_PRAUC_hires}, \textit{B-B-B} is no longer as strongly dominated by other input channel configurations as it was the case at lower input resolutions due to the blurring effect now being less severe.

\begin{table}[!ht]
\scriptsize
\begin{center}
\begin{tabular}{c|c|}
\cline{2-2}
\multicolumn{1}{l|}{}                   & \multicolumn{1}{l|}{\textbf{preprocessing}} \\ \hline
\multicolumn{1}{|c||}{\textbf{channels}} & \textbf{inc\_d}                             \\ \hline\hline
\multicolumn{1}{|c||}{\textbf{B-B-B}}    & 0.7715                                      \\ \hline
\multicolumn{1}{|c||}{\textbf{B-H-N}}    & \textbf{0.7736}                             \\ \hline
\multicolumn{1}{|c||}{\textbf{B-N-H}}    & 0.7718                                      \\ \hline
\multicolumn{1}{|c||}{\textbf{H-B-N}}    & 0.7676                                      \\ \hline
\multicolumn{1}{|c||}{\textbf{H-H-H}}    & 0.7681                                      \\ \hline
\multicolumn{1}{|c||}{\textbf{H-N-B}}    & 0.7722                                      \\ \hline
\multicolumn{1}{|c||}{\textbf{N-B-H}}    & 0.7695                                      \\ \hline
\multicolumn{1}{|c||}{\textbf{N-H-B}}    & 0.7687                                      \\ \hline
\multicolumn{1}{|c||}{\textbf{N-N-N}}    & 0.7696                                      \\ \hline\hline
\multicolumn{1}{|c||}{$\mu$}             & 0.7703                                      \\ \hline
\multicolumn{1}{|c||}{$\max(\Delta)$}    & 0.0060                                      \\ \hline
\end{tabular}
\caption{\textit{PRAUC} for $928 \times 384$ input resolution, BCE loss, and various combinations of input channel configurations. The highest \textit{PRAUC} in \textbf{bold}.}
\label{tab:transform_channel_PRAUC_vhires}
\end{center}
\end{table}

In out final experiment, we have investigated increasing the input resolution to $928 \times 384$. As shown in Table~\ref{tab:transform_channel_PRAUC_vhires}, this has resulted in further improvement in terms of \textit{PRAUC} reaching 0.0280 on average. It is also apparent that with the increase in the input resolution, the importance of interpolation diminishes -- the maximum difference among all the interpolation methods is 0.0060, albeit achieved at significantly increased computational cost.

\subsection{Error analysis}

In Figure~\ref{fig:radar} we depict the \textit{PRAUC} of our best model from Table~\ref{tab:transform_channel_PRAUC_vhires}, broken down by class/descriptor. As it can be seen, some descriptors seem to be relatively easy to classify; \textit{D01: Bar}, \textit{D02: Circular}, \textit{D04: 4 sided}, \textit{D07: Complex} and \textit{D08: Zigzag} all have $PRAUC > 0.9$, which is to be expected as these descriptors are relatively clear cut (see Table~\ref{tab:desciptor_code_names}). At the other end of the spectrum, \textit{D05: 5 sided} followed by \textit{D13: Hollow} and \textit{D01-02: Curved-wavy} are the most challenging. Note, that \textit{D05} is not only the least frequent in the dataset as per Table~\ref{fig:class_hist} (we've counteracted this by using class-weighting in the \textit{BCE} loss), but it is also one of the subtler descriptors in general. As it can be seen in Table~\ref{tab:desciptor_code_names}, \textit{D05} can for example be a rectangle with one of the corners `cut off', hence easy to confuse with \textit{D04: 4 sided}.  In a similar vein, \textit{D13: Hollow} can easily be confused with a circle (\textit{D02: Circular}), triangle (\textit{D03: 3 sided}), square/rectangle (\textit{D04: 4 sided}) etc. Some shapes on an impression can also represent multiple descriptors, for example \textit{D09: Text} and \textit{D10: Logo} will often overlap. Some overlays may be more complex such as \textit{D03: 3 sided} and \textit{D13: Hollow}. There may even be some examples of nested overlap such as \textit{D02-01: Target} (which implies \textit{D02: Circular}) and \textit{D13: Hollow}, as a circle with the centre missing is both a target and hollow. \textit{D14: Plain} is unusual in that when it applies to part of the shoe, it excludes the other descriptors from that area. 

\begin{table}[ht]
\begin{minipage}[b]{0.33\linewidth}\centering
\scriptsize
    \begin{tabular}{|l|c|}
    \hline
    \multicolumn{1}{|l|}{\textbf{Descriptor}} & \multicolumn{1}{l|}{\textbf{PRAUC}} \\ \hline
    \textbf{D01}                              & 0.9541                              \\ \hline
    \textbf{D01-01}                           & 0.7172                              \\ \hline
    \textbf{D01-02}                           & 0.6267                              \\ \hline
    \textbf{D02}                              & 0.9145                              \\ \hline
    \textbf{D02-01}                           & 0.7856                              \\ \hline
    \textbf{D03}                              & 0.7713                              \\ \hline
    \textbf{D04}                              & 0.9314                              \\ \hline
    \textbf{D05}                              & 0.4086                              \\ \hline
    \textbf{D06}                              & 0.7151                              \\ \hline
    \textbf{D07}                              & 0.9129                              \\ \hline
    \textbf{D08}                              & 0.9575                              \\ \hline
    \textbf{D09}                              & 0.7859                              \\ \hline
    \textbf{D10}                              & 0.7664                              \\ \hline
    \textbf{D11}                              & 0.8665                              \\ \hline
    \textbf{D12}                              & 0.7315                              \\ \hline
    \textbf{D13}                              & 0.5964                              \\ \hline
    \textbf{D14}                              & 0.7097                              \\ \hline
    \end{tabular}
    \begin{description}\item[\quad]\end{description} 
\end{minipage}\hfill
\begin{minipage}[b]{0.66\linewidth}\centering
\includegraphics[width=78mm]{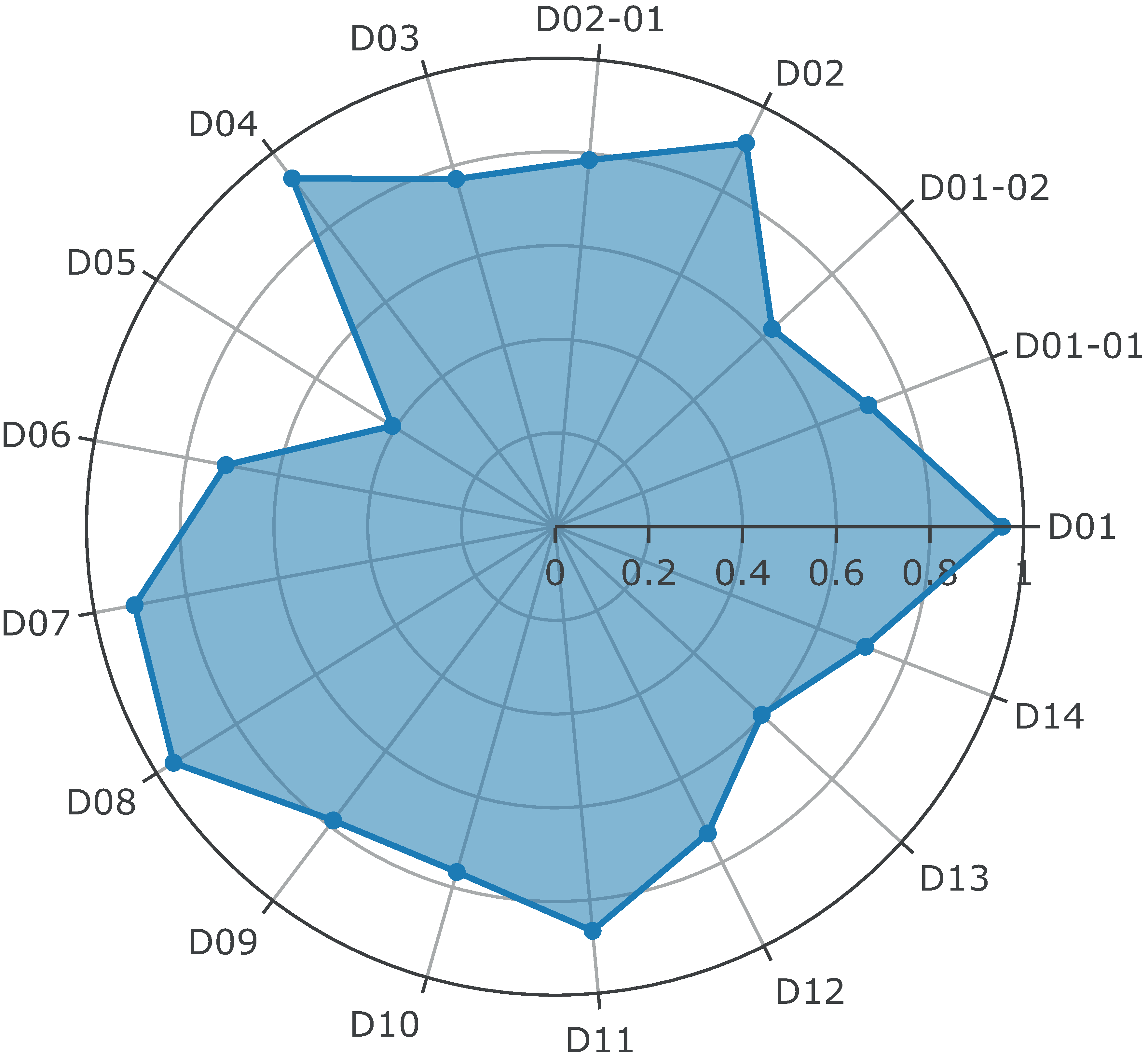}
\captionof{figure}{Per class PRAUC}
\label{fig:radar}
\end{minipage}
\end{table}

\begin{figure}[h]
    \centering
    \vspace*{-15mm}
    \includegraphics[width=1.2\linewidth]{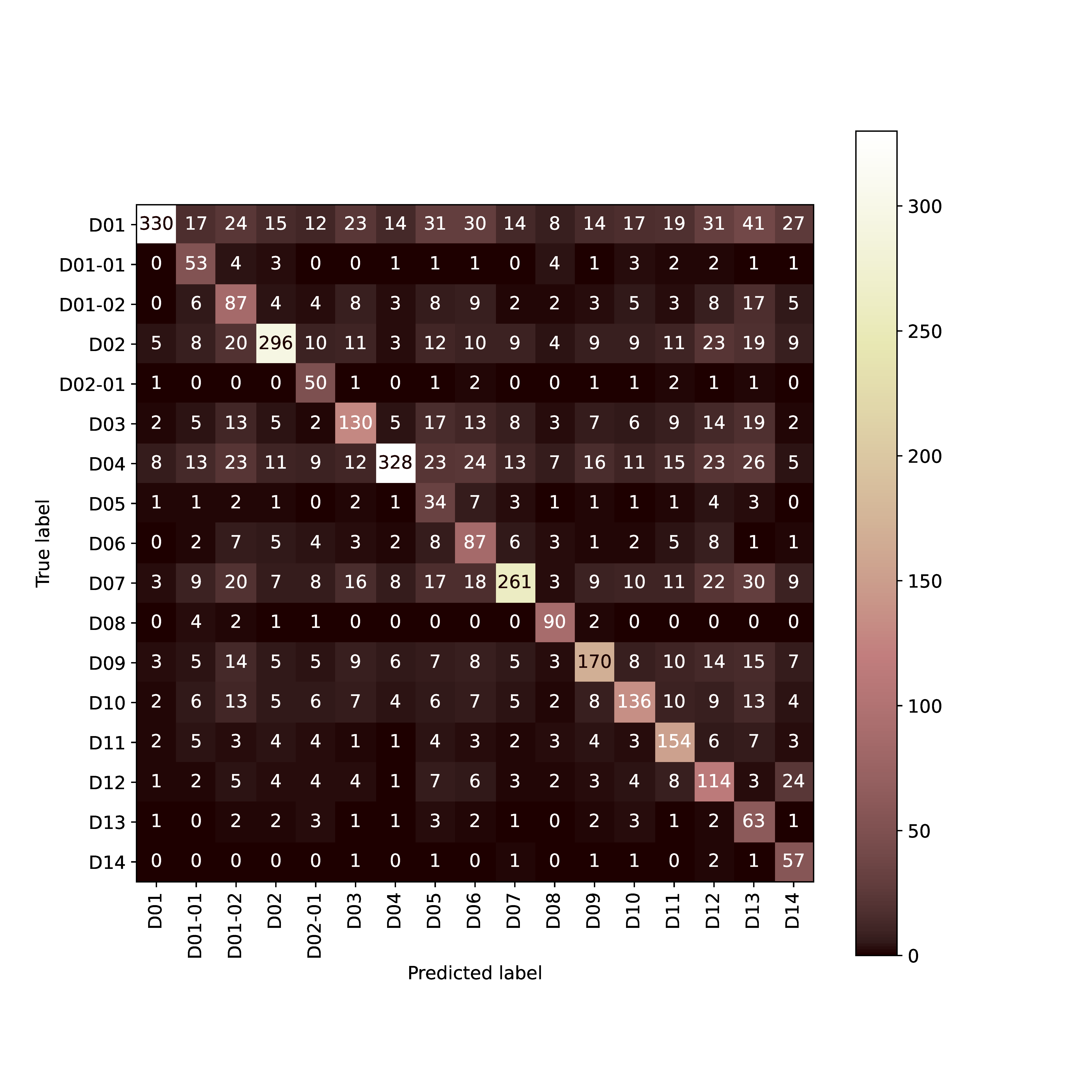}
    \vspace*{-20mm}
    \caption{Confusion matrix}
    \label{fig:confusion}
\end{figure}

In order to investigate this issue further, in Figure~\ref{fig:confusion} we show the confusion matrix generated for the validation dataset. For each validation image, if the predicted score for a descriptor which is not present in the image (false positive) exceeds the score for a descriptor which is in the image (true positive), then the two are considered confused. An example is given in Table~\ref{tab:confusion_example}. The actual labels are \textit{D02}, \textit{D05} and \textit{D06}. Since the score for \textit{D02} is the highest, 1 would be added to the diagonal entry for this descriptor. However, as the score for \textit{D04} (false positive) is higher than that for \textit{D05} and \textit{D06}, these are considered confused (i.e. either or both of \textit{D05} and \textit{D06} are classified as \textit{D04}) and hence $1/2$ (i.e. one over the number of potentially misclassified descriptors) is added to the entries \textit{D05-D04} and \textit{D06-D04} of the confusion matrix.

\begin{table}[!h]
\centering
\scriptsize
\begin{tabular}{|r|c|c|c|c|c|c|}
\hline
\textbf{Descriptor} & \textbf{D01} & \textbf{D02} & \textbf{D03} & \textbf{D04} & \textbf{D05} & \textbf{D06} \\ \hline
\textbf{Label}      & 0            & 1            & 0            & 0            & 1            & 1            \\ \hline
\textbf{Score}      & 0.3          & 0.9          & 0.2          & 0.8          & 0.6          & 0.6          \\ \hline
\end{tabular}
\caption{Example prediction to illustrate confusion matrix calculation}
\label{tab:confusion_example}
\end{table}

\begin{figure}[h]
    \centering
    \subfigure[$D01,D06 \rightarrow D13$]{\label{fig:D01asD13}\includegraphics[height=80mm]{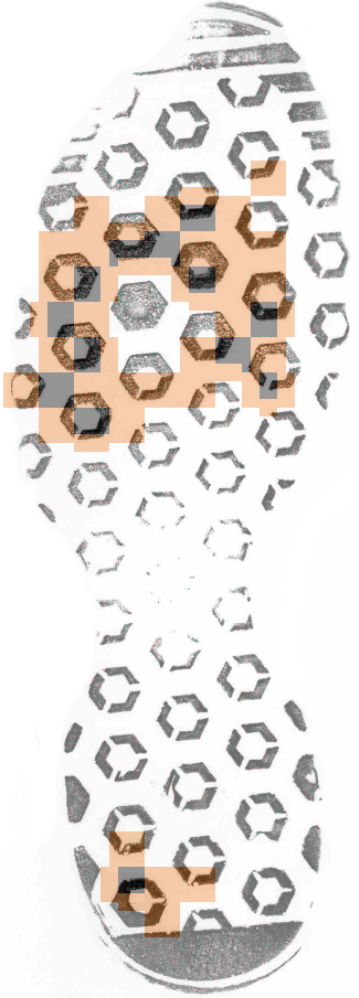}}
    \subfigure[$D01,D11 \rightarrow D04$]{\label{fig:D11asD04}\includegraphics[height=80mm]{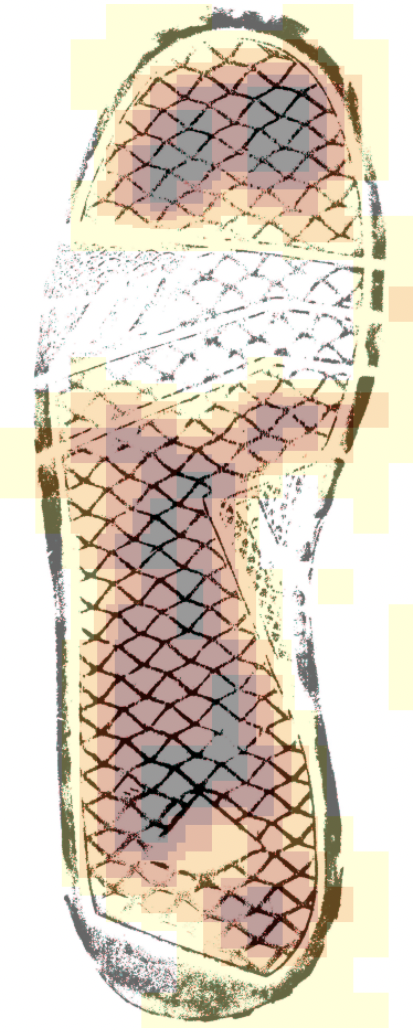}}
    \subfigure[$D05     \rightarrow D06$]{\label{fig:D05asD06}\includegraphics[height=80mm]{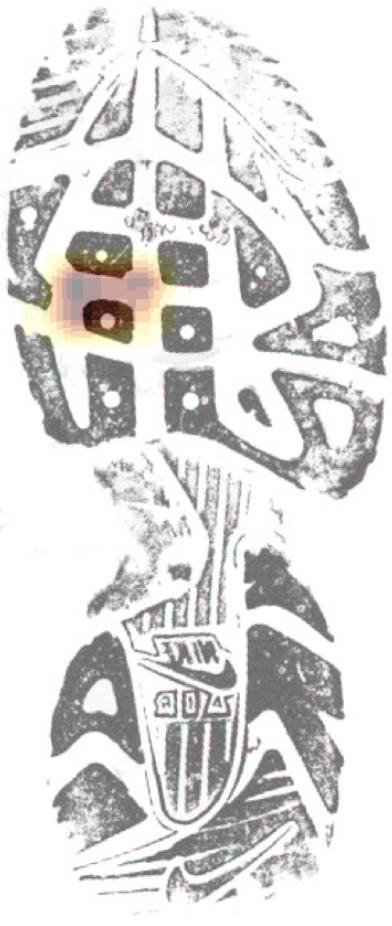}}
    \caption{High loss misclassification examples}
    \label{fig:high_loss_misclasf}
\end{figure}

As it can be seen in Figure~\ref{fig:confusion}, \textit{D01} is the most frequently misclassified descriptor which can be partially explained by its prevalence in the dataset. In Figure~\ref{fig:D01asD13} we show an example of a \textit{D01: Bar} in the top right corner of the print, which has not been detected by our model. At the same time, the model detected \textit{D13: Hollow} in the locations shaded in orange, although this particular shoeprint impression has not been labelled with \textit{D13} by the human expert. However, due to wear, some of the 6 sided shapes (\textit{D06}) closed, and indeed now fit the description of \textit{D13}. Another example of undetected \textit{D01} is given in Figure~\ref{fig:D11asD04}, where in addition \textit{D11: Lattice} has been misclassified as \textit{D04: 4 sided} in the areas highlighted by the heatmap -- the lattice indeed consists of 4 sided `cells'. In both of these examples, it is actually difficult to state which descriptor \textit{D01} was confused with; it appears that \textit{D01} was simply not detected, yet this would still be recorded in the confusion matrix due to the way in which the matrix was derived.

Another descriptor worth looking at is \textit{D05}, which as mentioned before is the most rare in the dataset and has the lowest \textit{PRAUC}. \textit{D05} is most often misclassified as \textit{D06} or \textit{D12}, while at the same time \textit{D03}, \textit{D04} and \textit{D07} are most often misclassified as \textit{D05}. An example can be seen in Figure~\ref{fig:D05asD06}.

It is worth noting, that all three examples of errors in Figure~\ref{fig:high_loss_misclasf} have been selected on the basis of the highest loss (i.e. they are as bad as it gets). It is reassuring that these do not result from the model behaving in an unexpected fashion, but are rather due to ambiguity in the inputs that may even cause disagreement between expert users, requiring resolving by expert panel.



\section{Conclusions}
The \textit{descriptor} identification task we have approached in this study is of great significance for the forensic practitioners in the UK and beyond. The \textit{descriptors} are an agreed standard for coding footwear patterns for different forces in the UK are in active use. Although a human performance benchmark is not available at this time, our model performs well with the \textit{PRAUC} of over $0.77$. The mistakes that the model tends to make are mostly justifiable, either by ambiguity or by overlaps in the input patterns, and are not unlike what an inexperienced human would make. The system that we have built has been deployed for testing by selected police forces.

In the process of building the model, we have experimented with a number of ways of feeding greyscale impressions to the ImageNet (RGB) pre-trained network. Our findings can be summarised as the following `best practices':

\begin{itemize}
    \item Preserve the aspect ratio of the input images. This seems particularly important if the object of interest (a shoeprint in our case) has aspect ratio significantly different than $1:1$ (i.e. `long and thin' or `short and fat'). This advice goes against the common practice in the computer vision community of `squishing' the input images to make them square in order to use ImageNet pre-trained models. This is unnecessary as current deep learning frameworks allow one to feed rectangular images to the ImageNet pre-trained models out of the box.
    \item Use as high input resolution as practical. In our experiments increasing the input resolution always led to higher \textit{PRAUC} albeit at the cost of significantly increased computations, which is an obvious constraint. The original resolution of the input images can also be a limitation as there's little point in upscaling such images.
    \item Use different interpolation methods to construct the three input channels from greyscale images. Although the effect of this approach that we have observed was modest, it was positive nevertheless. It also seems that using the Nearest Neighbour interpolation as one of the input channels is beneficial, while using three identical channels obtained via Bilinear interpolation is detrimental, particularly at lower input resolutions.
    \item Use a learnable preprocessing layer. In our experiments, these additional computations played a crucial role in the process of adapting greyscale inputs to be used with a colour-image pre-trained network, regardless of the interpolation method used. Learnable preprocessing combined with different interpolation methods to construct the three input channels gave the best results.
\end{itemize}

\section*{Acknowledgements}
The financial support of Innovate UK is acknowledged via the Knowledge Transfer Partners programme.

\bibliography{mybibfile}

\end{document}